\documentclass[10pt,twocolumn,letterpaper]{article}

\usepackage{cvpr}      

\usepackage[dvipsnames]{xcolor}

\definecolor{cvprblue}{rgb}{0.21,0.49,0.74}
\usepackage[pagebackref,breaklinks,colorlinks,citecolor=cvprblue]{hyperref}

\def\paperID{14642} 
\def\confName{CVPR}
\def\confYear{2024}

\usepackage{hyperref}
\usepackage{url}
\usepackage{booktabs}       
\usepackage{amsfonts}       
\usepackage{nicefrac}       
\usepackage{microtype}      
\usepackage{wrapfig}
\usepackage{latexsym}
\usepackage{amsfonts}
\usepackage{xcolor}
\usepackage{amsmath}
\usepackage{siunitx}
\usepackage{dcolumn}
\usepackage{arydshln}
\usepackage{multirow}
\usepackage{booktabs}
\usepackage{graphicx}
\usepackage{algorithm, algpseudocode}
\usepackage{graphicx}
\usepackage{verbatim}
\usepackage{stmaryrd}
\usepackage{trimclip}
\usepackage{subcaption}
\usepackage{url}
\usepackage{xspace}
\usepackage{colortbl}
\usepackage{color}
\usepackage[normalem]{ulem}
\usepackage[most]{tcolorbox}
\usepackage{tabularx}
\usepackage{xcolor}
\usepackage{fontawesome5}
\usepackage{amssymb}
\usepackage{pifont}

\definecolor{yellow}{RGB}{254, 242, 204}
\definecolor{blue}{RGB}{164, 194, 244}

\newcommand{\RN}[1]{%
	\textup{\lowercase\expandafter{\it \romannumeral#1}}%
}

\title{TRINS: Towards Multimodal Language Models that Can Read}
\author{
 ~
Ruiyi Zhang $^1$,
 ~
Yanzhe Zhang $^2$,
 ~
Jian Chen $^3$,
 ~
Yufan Zhou $^1$, 
 ~
Jiuxiang Gu $^1$,
\\
 ~
Changyou Chen $^3$,
 ~
Tong Sun $^1$ 
\\
 ~
$^1$ Adobe Research~~~~
$^2$ Georgia Institute of Technology~~~~ 
$^3$ State University of New York at Buffalo\\
 ~
{\small \texttt{\{ruizhang\}@adobe.com}}
}
\begin{document}
\maketitle
\begin{abstract}
Large multimodal language models have shown remarkable proficiency in understanding and editing images. However, a majority of these visually-tuned models struggle to comprehend the textual content embedded in images, primarily due to the limitation of training data.
In this work, we introduce TRINS: a Text-Rich image\footnote{In this work, we use the phrase ``text-rich images'' to describe images with rich textual information, such as posters and book covers.} INStruction dataset, with the objective of enhancing the reading ability of the multimodal large language model. 
TRINS is built upon LAION~\footnote{Work done during Q3 2023.} using hybrid data annotation strategies that include machine-assisted and human-assisted annotation processes. It contains 39,153 text-rich images, captions, and 102,437 questions.  Specifically, we show that the number of words per annotation in TRINS is significantly longer than that of related datasets, providing new challenges. Furthermore, we introduce a simple and effective architecture, called a Language-vision Reading Assistant (LaRA), which is good at understanding textual content within images. LaRA outperforms existing state-of-the-art multimodal large language models on the TRINS dataset, as well as other classical benchmarks. Lastly, we conducted a comprehensive evaluation with TRINS on various text-rich image understanding and generation tasks, demonstrating its effectiveness.
\end{abstract}
    
\section{Introduction}
\label{sec:intro}

Instruction tuning \citep{ouyang2022training, chung2022scaling} has shown a great generalization ability on unseen tasks and has contributed to the growing popularity of large language models (LLMs), such as ChatGPT~\cite{openai2023gpt4}. Recently, multimodal language models benefit from visual instruction finetuning~\citep{liu2023visual, li2023otter, Li2023LargeMM,zhu2023minigpt4,alayrac2022flamingo}, and have shown great success in real-world applications. These models leverage visual encoders such as CLIP-ViT \citep{dosovitskiy2020image, radford2021learning} to empower LLMs with image comprehension ability. However, challenges arise in comprehension of textual information within images, which may stem from the prevalence of natural images in training datasets, such as Conceptual Captions \citep{changpinyo2021conceptual} and COCO \citep{lin2015microsoft}), as highlighted by \citet{liu2023hidden}. Recognizing the importance of visual textual understanding for effective collaboration between agents and humans, \citet{zhang2023llavar} proposed enhancing end-to-end visual instruction-tuned models by introducing noisy Optical Character Recognition (OCR) annotations to improve vision-language alignment. In this work, we surpass existing achievements and collect a new Text-Rich image INStruction dataset named \textbf{TRINS}, which contains 39,153 text-rich images, captions, and 102,437 questions. 

TRINS is created in a semi-automatic manner for a more controllable and faithful collection. Specifically, we exploited large-scale pre-trained models such as CLIP~\cite{radford2021CLIP} and GPT-4 in the annotation process. This semi-automatic process significantly reduces the time and resources required for manual annotation and surprisingly improves the overall quality of annotations.
TRINS dataset is composed of three datasets for captioning, visual question answering (VQA), and image generation, respectively. 
Specifically, human-annotated captions for text-rich images are first collected because they can best translate text-rich images into texts. During this process, extracted OCR words and recognize-anything model tags are provided to the annotators for better and efficient annotations. With detailed image descriptions, VAQ data is built and fulfilled by large language models, such as GPT-4~\cite{openai2023gpt4} and LLaMA-2~\cite{touvron2023llama2}.   
In detailed statistics and analysis, we found that both annotated captions and collected question-answer pairs are more comprehensive and contain significantly more details than the existing dataset. Therefore, we show the superior advantage of TRINS compared to existing instruction fine-tuning datasets. 
As a by-product, high-quality image-caption pairs can serve as a good benchmark for text-rich image generation, which is still a very challenging task~\cite{chen2023textdiffuser}. At the same time, we propose a new, simple, and effective multimodal language model architecture that includes OCR as a component. We call it Language-vision Reading Assistant (LaRA) and show that LaRA fine-tuned on TRINS brings the best text-rich image understanding ability. Our contributions are as follows:
\begin{itemize}
    \item We introduce a novel dataset (TRINS) containing 39,153 captions and 102,437 high-quality text-rich image instruction pairs. TRINS is annotated with a novel semi-automatic annotation framework that is scalable and reliable. 
    \item We develop several evaluation benchmarks for text-rich image understanding and generation tasks. Various methods are evaluated on TRINS, demonstrating the effectiveness of the dataset.
    \item The TRINS datasets are high-quality and comprehensive, which is reflected not only in the dataset statistics but also from the results of multiple baseline models. Especially, our proposed LaRA finetuned on TRINS outperforms existing state-of-the-art methods on text-rich image understanding tasks.
\end{itemize}

\section{Related Work}
\paragraph{Multimodal Instruction Tuning}
Multi-modal instruction tuning, including image, video \citep{zhang2023video, maaz2023videochatgpt}, and audio \citep{Huang2023AudioGPTUA,zhang2023speechgpt} settings, has been an active research topic. MiniGPT-4 \citep{zhu2023minigpt4} uses ChatGPT to generate high-quality instruction-following data, while LLaVA \citep{liu2023visual} generates such data by prompting GPT-4 with captions and bounding boxes. LLaMA-Adapter \citep{zhang2023llamaadapter, gao2023llamaadapterv2} aligns text-image features using COCO data, and mPLUG-owl \citep{ye2023mplugowl} combines extensive image-text pairs for pretraining and a mixture of data for finetuning. Despite this, many models, according to \citet{liu2023hidden}, struggle with OCR tasks. InstructBLIP \citep{dai2023instructblip} addresses this by transforming 13 vision language tasks into an instruction-following format. mPLUGOwl~\cite{ye2023ureader,ye2023mplugowl} apply multitask instruction funetuing using existing document datasets. A comprehensive survey is available in \citet{li2023multimodal}. LLaVAR~\cite{zhang2023llavar} exploits GPT-4 to collect fine-tuning data without human annotations using OCR and captioning tools. It discovered that resolution plays a significant role in recognizing textual information and explored several options. Monkey~\cite{li2023monkey} performed a surgery between simple text labels and high input resolution, enabling remarkable performance in visually-rich document images with dense text. TRINS exploits human-machine collaboration for data annotation and can provide more accurate information, reducing the problem of hallucination. 
\paragraph{Text-Rich Image Datasets} 
Visual question answering or captioning datasets are widely used in task-specific fine-tuning and large multimodal model evaluation. 
TextCap~\cite{sidorov2020textcaps} is the first text-rich image captioning dataset. Compared to TextCap, TRINS-Cap provides more detailed annotations that can fulfill the requirement of instruction finetuing.
Text-OCR~\cite{textocr} aims to comprehend text in the context of an image, which is similar to our motivation but focuses more on text recognition in images instead of understanding. ST-VQA~\cite{STVQA} uses spatial and textual information to answer visually grounded questions, effectively integrating visual and textual cues. OCR-VQA~\cite{mishra2019ocrvqa} focuses on incorporating optical character recognition (OCR) into visual question answering (VQA), which operates primarily on text within images. TextVQA~\cite{textvqa} also takes advantage of the textual information present in the images to answer questions, but with an emphasis on open questions. DocVQA takes this one step further by applying VQA to document images, handling a variety of layouts and formats. InfoVQA~\cite{mathew2022infographicvqa} and ChartQA~\cite{masry2022chartqa} focus on specific subdomains and aim to answer questions about information graphics and chart images, respectively. In summary, these related works provide datasets for leveraging spatial and textual cues. TRINS-VQA is a dataset that exploits the semi-automatic annotation process. It can be used for general domain instruction finetuning and model evaluations. 
\begin{table}[t]
\small
\centering
\begingroup
\setlength{\tabcolsep}{8pt} 
\small
\begin{tabular}{l|l|l|l}
    \toprule
    \textbf{Dataset} & \textbf{Year}& \textbf{Size} & \textbf{Annotation Type}  \\ \hline
    OCR-VQA & 2019 & 200K & QA (\faRobot )  \\ 
    TextVQA & 2019 & 45K & QA (\faUsers )   \\ 
    TextCap & 2020 & 140k & Caption (\faUsers)  \\
    TextOCR & 2021 & 145k & Text Bbox (\faUsers)  \\
    DocVQA & 2020 & 50k & QA (\faUsers) \\ \hline
    TRINS (\textbf{Ours}) & 2023 & \begin{tabular}[l]{@{}l@{}}40k\\
    100k\\
    40k
    \end{tabular}  & \begin{tabular}[l]{@{}l@{}}Caption (\faUsers)\\
    QA (\faRobot~+ \faUsers)\\
    Text Bbox (\faRobot)
    \end{tabular}    \\
    \bottomrule
\end{tabular}
\endgroup
\caption{Comparison between TRINS and other related datasets.}
\vspace{-1em}
\label{tab: datasetreview}
\end{table}
\section{Text-Rich Image Instruction Dataset}

\begin{figure*}[htp]
    \centering
    \includegraphics[width=\textwidth]{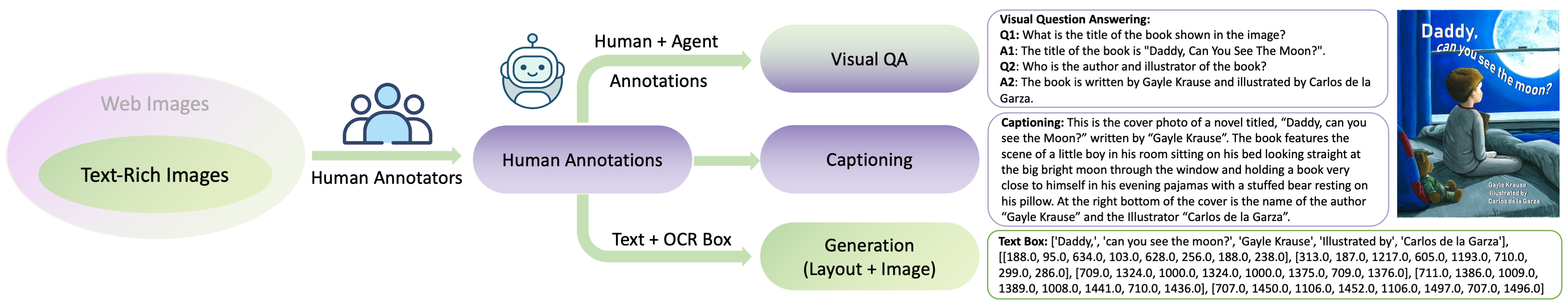}
    \caption{Overview of the TRINS data collection process, which consists of three datasets. Text-rich images are first selected from web images and then ask annotators to describe the image in detail. \RN{1}) TRINS-Cap is extracted from human annotations with heuristic data processing for text-rich image captioning tasks. \RN{2}) TRINS-VQA is built upon human annotations and generates question-answer pairs for training by prompting text-only LLMs. \RN{3}) TRINS-Gen combined human annotations and text boxes for text-rich image generation.}
    \label{fig:pipeline}
\end{figure*}

\begin{figure*}[htp]
    \centering
    \hspace{-4mm}
    \includegraphics[width=0.64\textwidth]{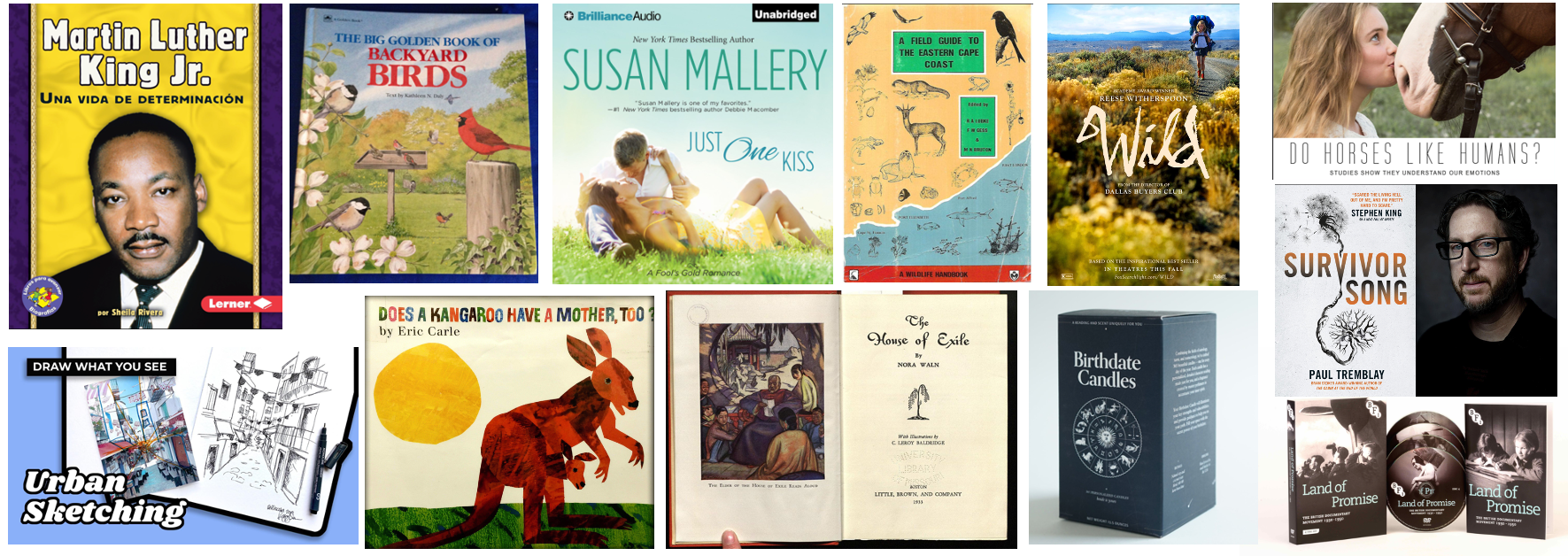}
    \includegraphics[width=0.34\textwidth]{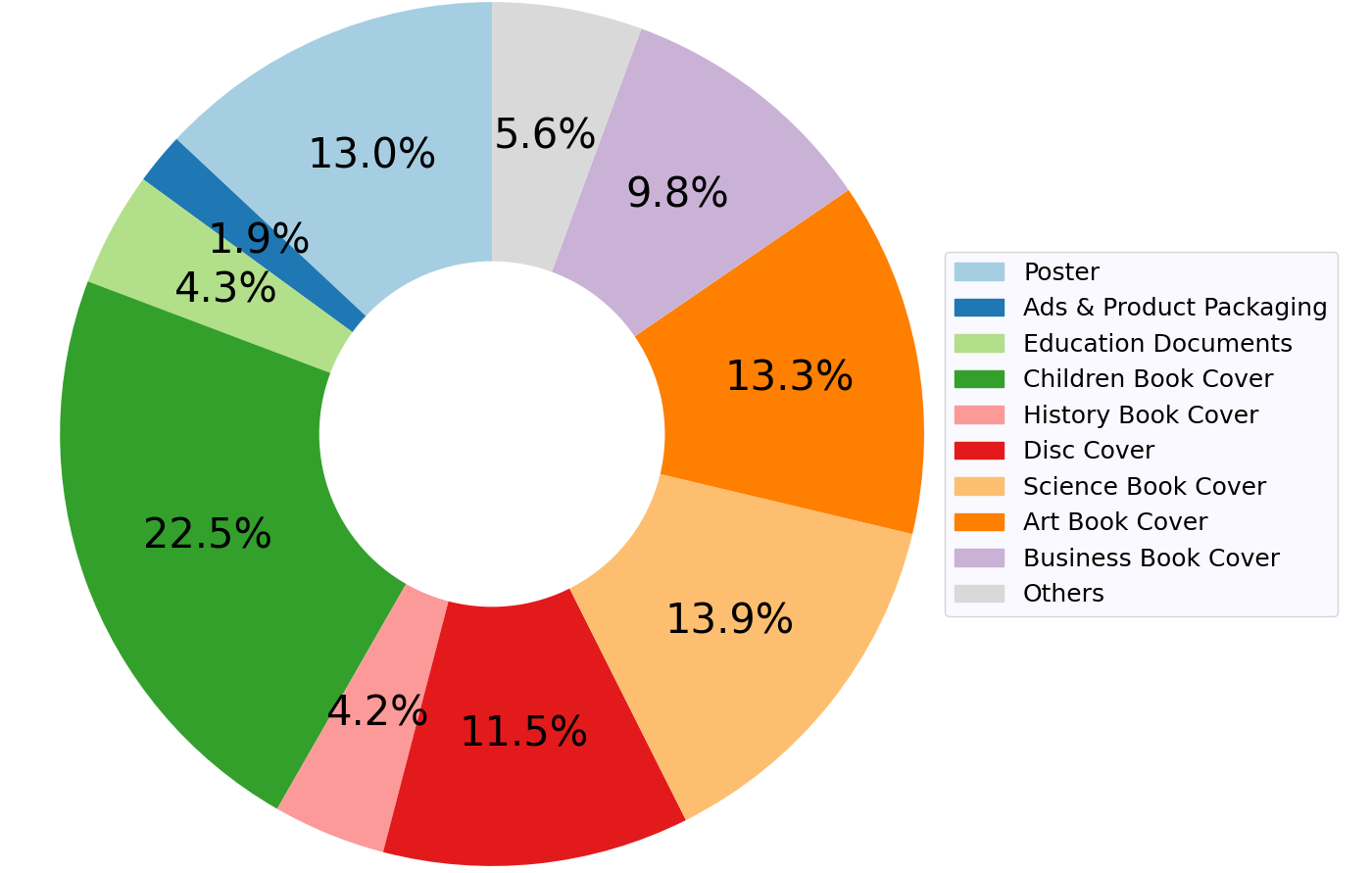}
    \caption{CLIP-based categorization of our collected images and selected representative data samples from each category.}
    \vspace{-1em}
    \label{fig:Data Collection}
\end{figure*}

To equip multimodal language models with the ability to recognize text and relate it to its visual context, we have curated a new dataset named Text-Rich Instruction (TRINS). The ultimate goal is to enable these models to have spatial, semantic, and visual reasoning between multiple text tokens and visual entities. In this section, we present TRINS, a dataset crafted through a semi-automatic process. Specifically, we leverage large-scale pre-trained models like CLIP~\citep{radford2021CLIP} and GPT-4 in the annotation process, offering potential advantages: (\RN{1}) Significant reduction in annotation time and resources:
using these models significantly reduces the time and resources required for manual annotation.
(\RN{2}) Enhancement of annotation data quality through post-processing: the involvement of large models contributes to improving the overall quality of annotation data through subsequent post-processing.
(\RN{3}) Functionality of large models as knowledge bases:
large models can serve as effective knowledge bases, aiding in the annotation process by virtue of their extensive training in diverse datasets.

We first outline the document image collection process for TRINS, utilizing CLIP models. Then, we present data statistics to facilitate a comprehensive understanding. We delve into three distinct tasks derived from the TRINS dataset in detail:
\RN{1}) TRINS-Cap: Visual Captioning,
\RN{2}) TRINS-VQA: Visual Question Answering and \RN{3}) TRINS-Gen: Text-to-Image Generation. The overview of the TRINS data collection process is illustrated in Figure \ref{fig:pipeline}. To provide a succinct overview, TRINS-Cap undergoes full annotation by human annotators, while TRINS-Gen and TRINS-VQA are constructed based on TRINS-Cap with the help of pre-trained models. A summary of the various TRINS datasets is presented in Table \ref{tab: stats} of Appendix \ref{app:datadetail}.
%
%
\subsection{Machine-Assisted Text-rich Image Selections} 
Beginning with the LAION-5B dataset\footnote{\url{https://huggingface.co/datasets/laion/laion-high-resolution}} \citep{schuhmann2022laion}, our objective is to selectively retain images that exhibit a significant presence of text. Recognizing that document images typically contain substantial textual content, we initially formed a binary classification dataset by combining natural images with document data. Subsequently, we trained an image classifier using a DiT \citep{2022DIT} base backbone, fine-tuned on the RVL-CDIP dataset \citep{harley2015evaluation}. The purpose of this classifier was to predict whether an image contains text. Then a subset was constructed by selecting images with a predicted probability greater than 0.8, while also adhering to the criteria $p(\text{watermark}) < 0.8$ and $p(\text{unsafe}) < 0.5$, where both probabilities are derived from the metadata of the LAION dataset. 
\begin{figure}[t]
    \centering
    \begin{subfigure}[b]{0.235\textwidth}
        \includegraphics[width=\textwidth]{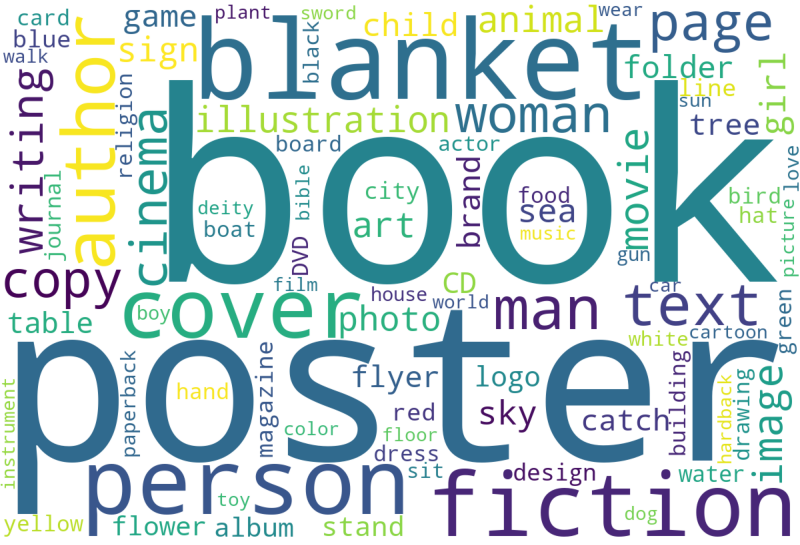}
        \caption{ }
        \label{fig:wordcloud1}
    \end{subfigure}
    \hfill
    \begin{subfigure}[b]{0.235\textwidth}
        \includegraphics[width=\textwidth]{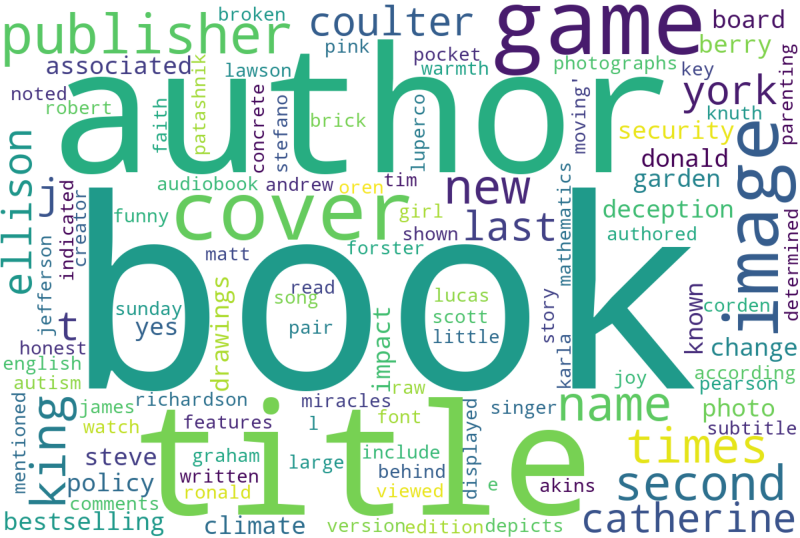}
        \caption{ }
        \label{fig:wordcloud2}
    \end{subfigure}
    \vspace{-2em}
    \caption{Word clouds of (a) predicted tags and (b) detected words from the text-rich images of TRINS.}
    \label{fig:images}
    \vspace{-2em}
\end{figure}
\begin{figure*}[t]
    \centering
    \begin{subfigure}[a]{0.24\textwidth}
        \includegraphics[width=\textwidth]{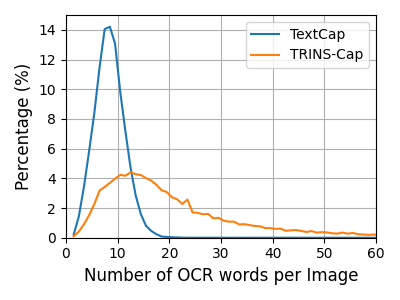}
        \vspace{-2em}
        \caption{ }
        \label{fig:stats1}
    \end{subfigure}
    \begin{subfigure}[a]{0.24\textwidth}
         \includegraphics[width=\textwidth]{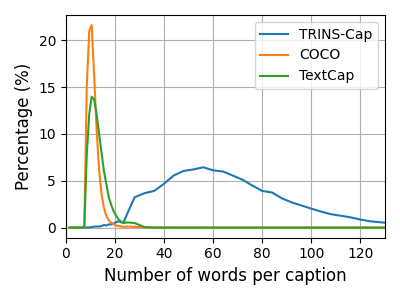}
        \vspace{-2em}
        \caption{ }
        \label{fig:stats2}
    \end{subfigure}
    \begin{subfigure}[a]{0.24\textwidth}
        \includegraphics[width=\textwidth]{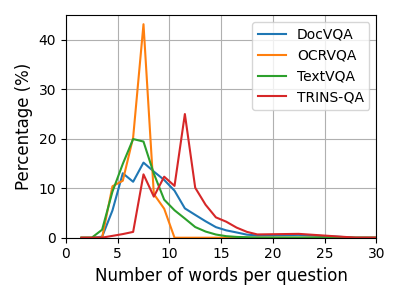}
        \vspace{-2em}
        \caption{ }
        \label{fig:stats3}
    \end{subfigure}
    \begin{subfigure}[a]{0.24\textwidth}
        \includegraphics[width=\textwidth]{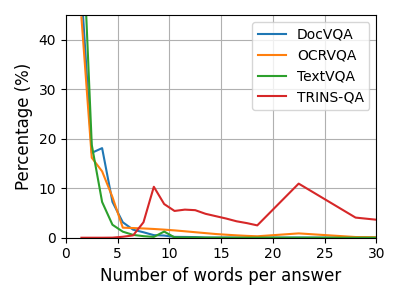}
        \vspace{-2em}
        \caption{ }
        \label{fig:stats4}
    \end{subfigure}
    \vspace{-1em}
    \caption{OCR word (a), Caption (b), Question (c) and Answer (d) statistics for TRINS.}
    \label{fig:stats}
    \vspace{-1.3em}
\end{figure*}
Acknowledging the noise introduced due to the classifier's limitations, we further refined the dataset by incorporating human judgment. A random sample of 20,000 images from the filtered LAION-5B was clustered into 50 groups based on CLIP-ViT-B/32 visual features. After inspecting the clustering results, one cluster was meticulously chosen, encompassing diverse text-rich images such as posters, covers, advertisements, and educational documents. This cluster model then served as the filtering mechanism for collecting images that comprise the TRINS dataset. 
For reference, we present a CLIP-based categorization \citep{radford2021learning} in Figure \ref{fig:Data Collection} to depict the distribution of images in the collected data. The major class is book cover images, further categorized on the basis of book themes and contents. To enhance our understanding of text-rich images, we employed a Recognize Anything Model (RAM) \citep{zhang2023recognize,huang2023tag2text} to extract tags from TRINS images. Figure \ref{fig:wordcloud1} displays word clouds of RAM tags, where ``book'' and ``poster'' emerge as major keywords. Additionally, we utilize the \href{https://azure.microsoft.com/en-us/updates/computer-vision-v3-preview-6/}{Azure Read API} and PaddleOCR to extract text within TRINS images. The word cloud of the extracted texts is presented in Figure \ref{fig:wordcloud2}. Figure \ref{fig:stats1} illustrates the distribution of OCR words per image, indicating that most of the images in TextCap have fewer than 10 words, while the TRINS images average 31.4 OCR words. Recognizing texts within TRINS images is more challenging because of the presence of numerous small words.
In summary, TRINS images encompass rich visual content, seamlessly integrating textual information into the image context.

\begin{figure}[htp]
    \centering
    \vspace{-0.2em}
    \hspace{-4mm}
    \includegraphics[width=0.45\textwidth]{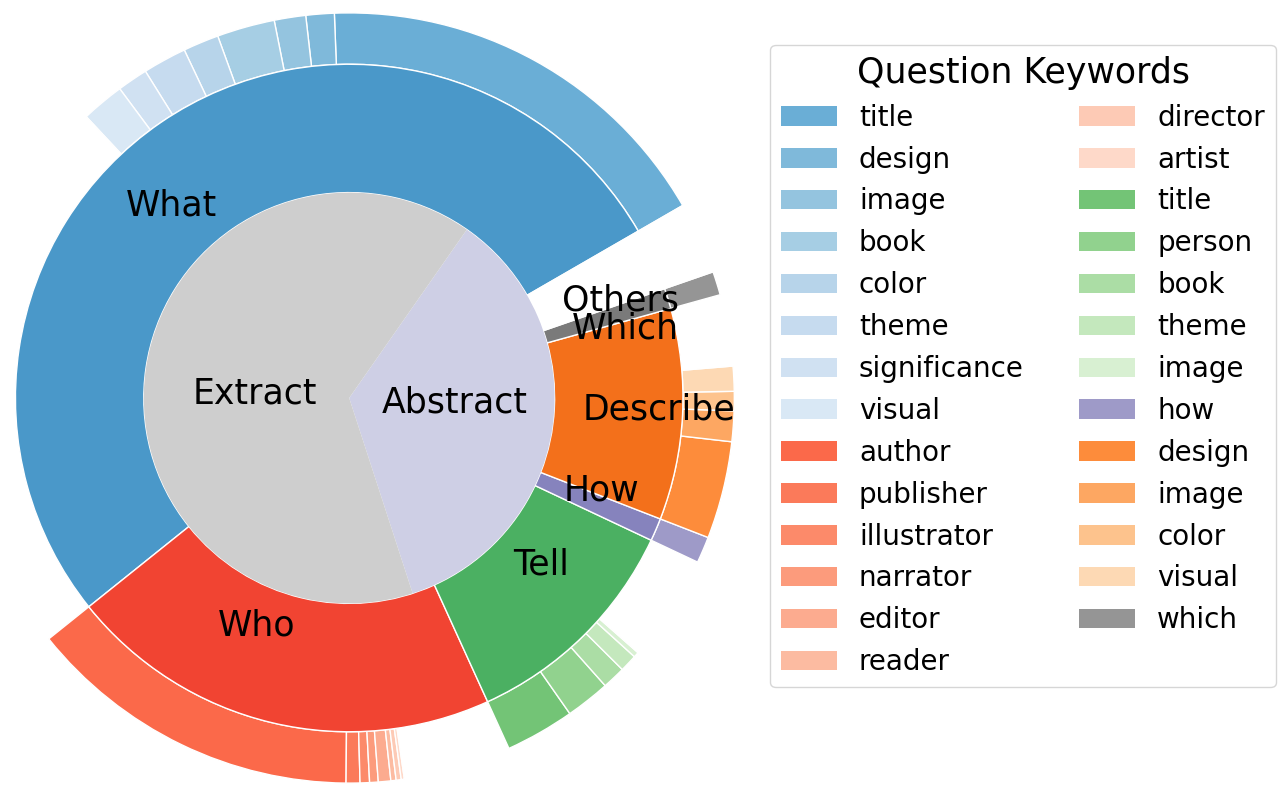}
    \caption{Question type statistics based on key words.}
    \vspace{-1em}
    \label{fig:TRINSqa}
\end{figure}

\subsection{Annotation Details} 
\textbf{Annotator Selections}\quad All annotators (with the tag \textit{100\% Job Success} and \textit{ Top Rated Plus}) are native English speakers and have experience with document annotations. We first asked all annotators to annotate a 200-image set and provided them with detailed annotation guidelines with multiple examples. In addition, we use Labelbox as an annotation tool and set quality control questions.\\
\textbf{Heuristic Filters}\quad We first use EasyOCR to extract texts from images and retain text phrases with more than three characters, a height greater than 5\% of the canvas height, and a confidence score greater than 0.1. For each phrase retrieved, we employ an edit distance-based string matching algorithm (due to potentially erroneous OCR results) to search for its optimal matching substrings within the human-generated caption. The average score for all extracted phrases serves as a metric.\\
\textbf{Manual Reviews} We accept annotations with high metric scores and reject the lowest for rework. We manually review other annotations.\\ 
\textbf{Sensitive Images}\quad  We combined neural models with human efforts to filter the images. The first step involves an initial data filtering by 2-3 individuals to filter out sensitive images for training. The second step involves hiring additional people to perform a further screening on the data. We engaged annotators from various countries to check the selected images. 

\subsection{TRINS-Cap: Text-Rich Image Captioning}
\label{dataset:cap}
\paragraph{Annotation Process} 
TRINS-Cap is a dataset fully annotated by human annotators. We hired 20 native English speakers with experience in document annotation through Upwork. The annotation process, conducted in LabelBox, involved a total of 2,079 hours to annotate 40,576 text-rich images, with an additional 159 hours allocated for result review. After filtering low-quality annotations and addressing missing images, we obtained a final set of 39,153 image-annotation pairs. The dataset is partitioned into train, validation, and test splits with sizes of 29,153, 5,000, and 5,000, respectively. All annotations undergo an initial automated review that involves matching the OCR words with the annotations. Subsequently, human evaluators conduct a thorough review, rejecting annotations with errors, and prompting annotators to rework them for enhancement.
We provide comprehensive annotation instructions to all annotators to ensure that each annotation includes:
(\RN{1}) detailed descriptions of visual components. (\RN{2}) describe texts' location, attributes, and put texts into annotations. (\RN{3}) optional insights or abstract descriptions.\\
\textbf{Statistics and Analysis}\quad
The primary objective of the annotation process is to facilitate a human or machine's full comprehension of the information conveyed in the image without direct viewing. Consequently, the average annotation length for TRINS is 65.1 words, significantly exceeding that of COCO (10.6 words) and TextCaps (12.4 words). Figure~\ref{fig:stats2} shows the caption length distributions for TRINS-Cap, COCO, and TextCap, demonstrating the comprehensive nature of the TRINS-Cap annotations. TRINS with more contexts can generally provide a better description of complex images, where short captions are insufficient. Hence, LLMs fine-tuned on TRINS can better understand images with complex texts and layouts, which has been further verified in Section \ref{sec:exp}.

\subsection{TRINS-VQA: Multimodal Question Answering}
\label{dataset:qa}
The annotation process for question answering is inherently complex, primarily due to the necessity for annotators to generate high-quality questions. Creating an effective question is more challenging than providing an answer. As a result, annotators frequently gravitate toward formulating concrete and extractive questions ({\em e.g.}, ``Who is the author of this book?'') rather than abstract ones ({\em e.g.}, ``How does the design of the book cover reflect the content of the book?'').
We introduce semi-automatic annotation methods to generate high-quality visual question-answering data for TRINS-VQA. This dataset is designed to train general vision language assistants through instruction fine-tuning, and its benefits on model performance are evaluated in Section~\ref{exp:vqa}. TRINS-Cap, on the other hand, serves as human-assisted annotations, offering a comprehensive but non-instructive dataset for fine-tuning. 
To utilize the wealth of high-quality annotations available, we incorporated semi-automatic annotation by using large language models (LLMs) such as OpenAI's GPT-4~\cite{openai2023gpt4} and Llama-70B~\cite{touvron2023llama,touvron2023llama2} to enhance our data annotation pipeline. OCR results and detailed descriptions of each image are provided to LLMs. Furthermore, high-quality human-crafted demonstrations and detailed annotation rules are provided to LLMs. One demonstration focused on extract questions, while the other emphasized abstract questions, creating a more balanced dataset.
\vspace{-1em}
\paragraph{Human Annotations} 
To facilitate a robust evaluation of model performance, we hired 10 Upwork annotators, whose native language is English, to annotate the test dataset, following a methodology similar to previous work~\cite{mathew2020docvqa,textvqa,STVQA}. The test dataset comprises 5,000 images with 18,764 question-answer pairs. These data collected are used exclusively for evaluation purposes.
\vspace{-1em}
\paragraph{Statistics and Analysis}
Building upon prior research \citep{wang2022selfinstruct, alpaca, liu2023visual}, we provide visualizations of instructions in Figure~\ref{fig:TRINSqa} based on question keywords. The inner cycle illustrates the distribution of the first word in the questions, while the outer cycle presents extracted keywords determined by carefully designed heuristics. Types of questions are categorized according to keywords found in questions.

In Figure~\ref{fig:stats3} and \ref{fig:stats4}, we present statistics on the number of words per question-answer pair, comparing them with previous work. Generally, the average length of questions for TRINS-VQA is 10.5, surpassing that of DocVQA (8.3), OCR-VQA (6.5), and TextVQA (7.1). Surprisingly, the average answer length for TRINS is 23.9, significantly longer than related datasets (all less than 4). This discrepancy arises from TRINS containing more abstract questions that typically have longer answers. 
Similarly, the dataset is divided into train, validation, and test splits. For extract questions, the accuracy of the answers is calculated similarly to ~\citet{liu2023visual}, while for abstract questions, generation metrics such as BLEU scores are used to assess the quality of the answers.

The question-answering and instruction data we obtain are extensive, encompassing a balanced mix of extract and abstract questions. This includes detailed descriptions, summaries, question-answer pairs, tasks that promote creativity and the generation of novel thoughts, and conversational tasks. The dataset spans a diverse range of concepts, ranging from visual presentation and visual language relations to intricate reasoning tasks. 
\begin{figure}[t!]
    \centering
    \hspace{-4mm}
    \includegraphics[width=0.4\textwidth]{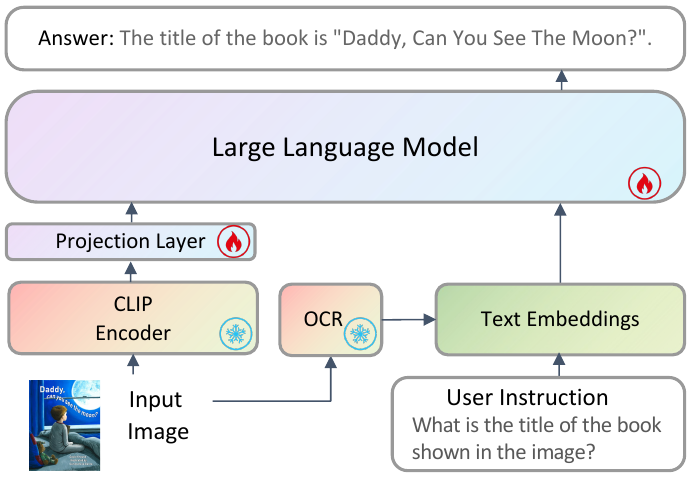}
    \caption{Model overview of the LaRA. The CLIP model processes the input image to generate patch-wise features. These features then serve as input to a projection layer, yielding visual tokens. Concurrently, an OCR tool extracts textual data from the image, which is then merged with the user instruction.}
    \vspace{-1em}
    \label{fig:baseline}
\end{figure}
Compared to previous methods~\cite{liu2023visual}, captions generated by models such as BLIP-2~\citep{li2023blip2} exhibit less informativeness than detailed human annotations, as demonstrated in Section~\ref{exp:cap}. Additionally, captioning models may be susceptible to hallucinations \citep{rohrbach2018object}, a concern mitigated in TRINS, which provides more comprehensive and reliable visual descriptions. Although OCR tools are robust, they can introduce noisy information. To address this, we utilize both the \href{https://learn.microsoft.com/en-us/azure/ai-services/computer-vision/overview-ocr#read-api}{Azure Read API} and \href{https://github.com/PaddlePaddle/PaddleOCR}{PaddleOCR} to extract text information. We added the potential unreliability in our system prompts to LLMs, instructing them to generate questions with assured answers. 
Ultimately, we directly leverage the responses from large language models (LLMs) to construct TRINS-VQA. The quality of the instruction data can be further enhanced through self-alignment~\citep{li2023self} or by seeking verification from human annotators. Although this has not been explored in this work, we leave it as a potential direction for future research.

\begin{figure*}[htp]
    \centering
    \hspace{-4mm}
    \includegraphics[width=0.95\textwidth]{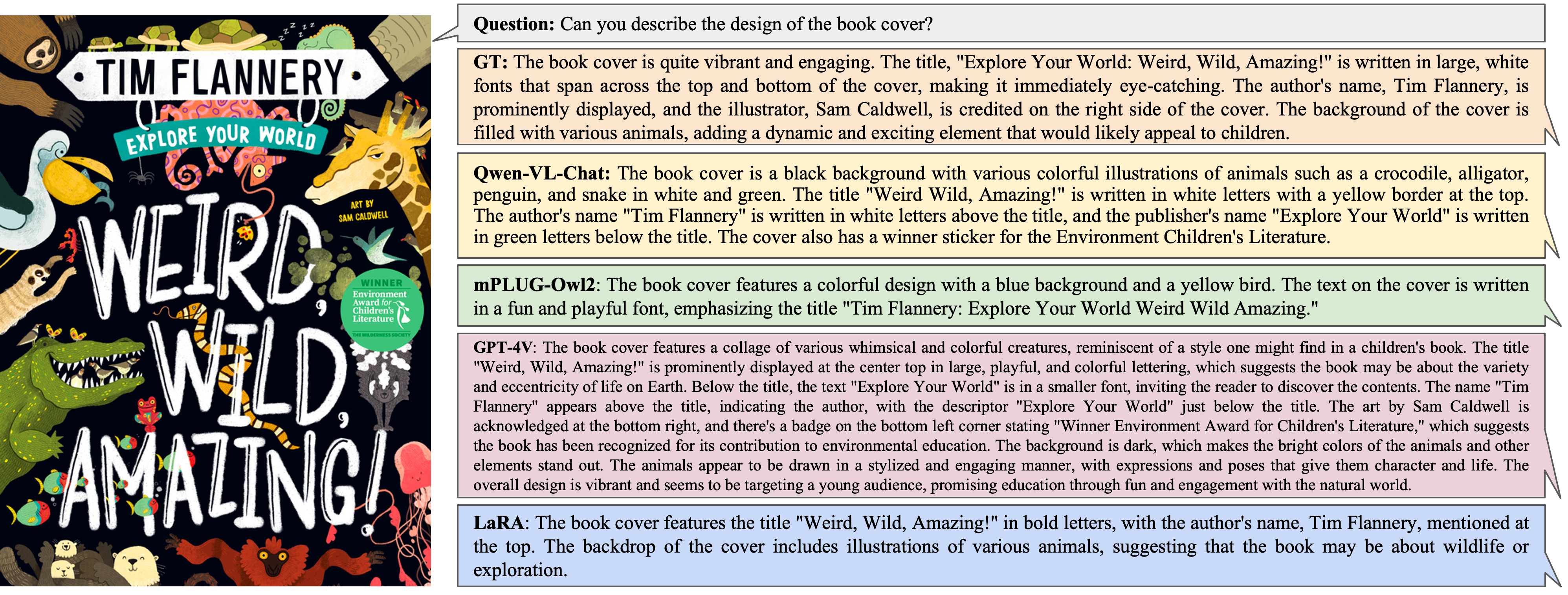}
    \vspace{-0.5em}
    \caption{Examples generated by different multimodal language models on the abstract TRINS-VQA benchamrk.}
    \vspace{-0.9em}
    \label{fig:qaexample}
\end{figure*}

\section{The Language-vision Reading Assistant (LaRA) Model}
To verify our data, we adopt a simple architecture based on LLaVA~\cite{liu2023visual} for language-vision reading assistant, as illustrated in Figure~\ref{fig:baseline}. A low-resolution visual encoder is a significant bottleneck for extracting textual information from images~\cite{zhang2023llavar}. On the contrary, a high-resolution visual encoder generates an excessive number of patches, drastically impacting model efficiency. Our conviction is that a low-resolution visual encoder is designed to capture visual information more effectively, including layouts. In contrast, a considerably smaller OCR tool is employed to extract text from high-resolution images. Instruction-tuning on TRINS could serve as an effective method of training LLMs to better align OCR texts and extract visual features. In addition, LLMs can autonomously rectify errors generated by OCR tools.
For the visual encoder $V$, we employ \texttt{CLIP-ViT-L/14-336} at a resolution of $336\times 336$. The grid features before the last transformer layer are then mapped into the word embedding space of the language decoder using a trainable projection matrix $W$. Regarding the language decoder $D$, we utilize Vicuna-1.5-13B \citep{vicuna2023}, a language model tuned for instructions through LLaMA 2~\citep{touvron2023llama}.

We follow the two-stage training design of LLaVA but adopt the pre-trained projection layer of LLaVAR~\cite{zhang2023llavar}. Training targets remain similar: generating \textbf{\textit{output responses}} (\emph{$<$res$>$}) for the \textbf{\textit{input instructions}} (\emph{$<$ins$>$}), alongside OCR results (\emph{$<$ocr$>$}). The transformed image tokens (\emph{$<$img$>$}) are introduced before or after the first input instruction randomly when building the instruction finetuning data. During the finetuning stage, both the projection matrix $W$ and the language decoder $D$ are trained. We consolidate our nearly 90K visual question-answering data with the 158K instruction-following data from LLaVA to form the training set. 
It should be noted that the visual encoder remains frozen throughout the training period. Compared with previous approaches, LaRA incorporates OCR words as part of the input, a simple way to enhance visual text understanding.

 \begin{table}[htp]
\centering
\small
\setlength{\tabcolsep}{3.5pt}
\begin{tabular}{c|ccccc|c}
\hline       Method
             & Recog. & VQA$^{S}$ & VQA$^{D}$ & KIE & Final Score \\ \hline
Gemini       & \textbf{215}                   & \textbf{174}                         & 128                   & 134               & \textbf{651}               \\ 
GPT-4v        & 167                   & 163                         & \textbf{146}                   & \textbf{160}              & 636               \\
\hline
Monkey       & 174                   & \textbf{161}                         & \textbf{91}                    & 88              & 514               \\
mPLUG-Owl2   & 153                   & 153                         & 41                    & 19                & 366               \\
LLaVAR      & 186                   & 122                         & 25                    & 13               & 346               \\
LLaVA1.5-13B & 176                   & 129                         & 19                    & 7                & 331               \\
mPLUG-Owl    & 172                   & 104                         & 18                    & 3                 & 297               \\
MiniGPT-V2   & 124                   & 29                          & 4                     & 0                & 157               \\
\hline
LaRA      & \textbf{211}                   & 147                         & 85                    & \textbf{105}               & \textbf{548}               \\
\hline
\end{tabular}
\caption{Results of LMMs on OCRBench. Recog. represents text recognition, VQA$^{S}$ represents Scene Text-Centric VQA, VQA$^{D}$ represents Document-Oriented VQA.}
\label{tab:OCRBench}
\vspace{-1.5em}
\end{table}

\section{Experiments}
\label{sec:exp}
In this section, we present three downstream tasks based on the TRINS dataset and outline their evaluation metrics. The proposed method LaRA is used in both the text-rich image summarization and visual question-answer tasks. All experiments were conducted on NVIDIA A100 80GB GPUs. In fine-tuning, we use a cosine annealing schedule with an initial learning rate of $2e^{-5}$ and a batch size of 32. 

\subsection{TRINS-VQA: Text-Rich Image Visual Question Answering}
\label{exp:vqa}

\begin{figure*}[htp]
    \centering
    \hspace{-4mm}
    \includegraphics[width=0.95\textwidth]{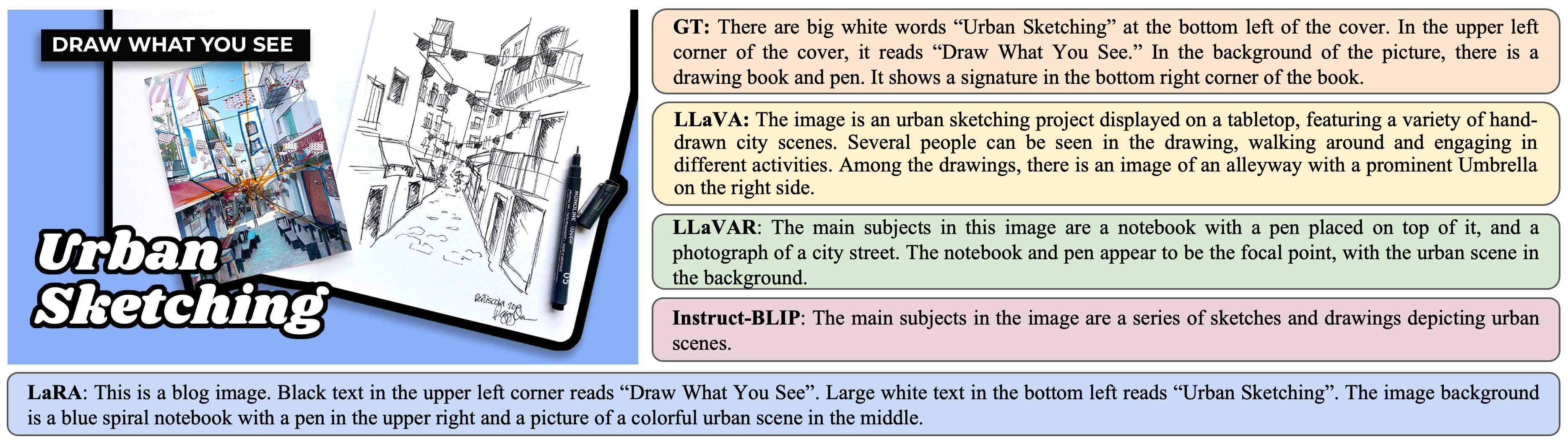}
    \vspace{-0.5em}
    \caption{Examples generated by different multimodal language models on the TRINS-Cap benchamrk.}
    \vspace{-0.5em}
    \label{fig:capexample}
\end{figure*}
\begin{table*}[htbp]
\small
\setlength{\tabcolsep}{7pt}
\centering
\begin{tabular}{l | c| c | ccccccc}
\toprule
\multirow{2}{*}{\textbf{Method}}  & \multirow{2}{*}{\textbf{Resolution}} & \textbf{Extract} &\multicolumn{7}{c|}{\textbf{Abstract}} \\
\cline{4-10}
 &   & \textbf{Accuracy} & \textbf{B@1} & \textbf{B@2} & \textbf{B@3} &\textbf{ B@4} & \textbf{METEOR} & \textbf{ROUGE} & \textbf{CIDEr} \\
\hline
Instruct-BLIP~\citep{Dai2023InstructBLIPTG} & \multirow{2}{*}{$224^2$}  & 43.9 & 13.5 & 9.2 & 6.7 & 5.2 & 8.9 & 16.7 & 23.5 \\
Mini-GPTv2 & & 15.3 & 29.4 & 19.4 & 13.8 & 10.4 &17.2 & 28.7 & 44.3 \\
mPLUG-Owl2~\citep{ye2023mplug} & \multirow{2}{*}{$448^2$} & 61.0 & 35.9 & 25.5 & 19.3 &15.2 & 17.7 & 38.0 & 97.2\\
Qwen-VL~\citep{bai2023qwen} & & \textbf{63.6} & 40.2 & 27.3 & 19.9 & 15.2 & 20.4 & 36.4 & 79.4 \\
LLaVA \citep{liu2023visual} & \multirow{4}{*}{$336^2$} & 23.7 & 33.1 & 20.9 & 14.4 & 10.5 & 18.0 & 31.8 & 48.2 \\
LLaVAR \citep{zhang2023llavar} & &51.7 &  38.0 & 25.9 & 19.1 & 14.7 & 17.4 & 35.6 & 84.9 \\
LLaVAR w/ OCR &  & 58.1 &40.3 & 28.0 & 20.9 & 16.3 & 18.7 & 37.3 & 97.3 \\

LLaVA 1.5~\cite{liu2023improved} & &38.8& 29.8  & 20.9  & 15.8  & 12.6  & 20.8  & 34.9 & 40.7   \\
\hline
{LLaVAR (finetuned)} & \multirow{2}{*}{$336^2$} & 61.2  & \textbf{45.3} & 31.9 & 23.9 & 18.7 & 20.4 & 39.6 & 104.7 \\
LaRA & & 62.8 & 45.1 & \textbf{34.2} &\textbf{ 27.3} &\textbf{ 22.6} & \textbf{21.9} & \textbf{46.5} & \textbf{186.6} \\
\bottomrule
\end{tabular}
\vspace{-0.7em}
\caption{Results of different models on TRINS-VQA for text-rich image question-answering tasks.}
\vspace{-1.1em}
\label{tab:TRINS-VQA}
\end{table*}

We first performed experiments to evaluate the zero-shot performance of LaRA on classical benchmarks~\cite{wu2023ocr,liu2023visual}. The results are reported in Table \ref{tab:OCRBench} and Table \ref{tab:vqa_result}. The proposed LaRA model exhibits significant performance improvement across all the datasets compared to other models. Even without OCR, LaRA outperforms other models in most cases, highlighting its robustness and effectiveness in handling visual question-answering tasks. The inclusion of OCR in LaRA further enhances performance, suggesting that Large Language Models (LLMs) can proficiently utilize textual information. However, the ability to directly extract text from images remains limited. LaRA, with its simple structure, significantly boosts model performance on text-rich images, offering an alternative solution to overcome the limitations of pre-trained image encoders. 
We further evaluate different methods on the TRINS-VQA dataset, as shown in Table \ref{tab:TRINS-VQA}. For extraction questions, we use the same metric as \citet{wu2023ocr}. For abstract questions, where the answer is typically a longer sentence, we evaluate them based on text similarity metrics such as BLEU~\citep{lin2004rouge}, ROUGE~\citep{lin2004rouge}, and CIDEr~\citep{vedantam2015cider}. In zero-shot inference, LLaVAR with OCR exhibits the best performance, reinforcing the importance of extracting textual information. 
Furthermore, mPLUG-Owl2 and Qwen-VL perform well and represent the best methods in extract question evaluations, showing that a high-resolution encoder can significantly improve model performance. Instruct-BLIP demonstrates good performance on extract questions, but did not fare as well on abstract questions, given that the answers provided are usually short and concise. Figure \ref{fig:qaexample} shows an example of the responses of different methods on the abstract TRINS-VQA dataset, and more examples can be found in the Appendix \ref{app:moreexamples}. Qwen-VL includes all details but does not provide high-level insights, such as the ground-truth annotation. Both mPLUG-Owl2 and GPT-4V suffer from hallucination issues.

\begin{table*}[htbp]
 \small
 \setlength{\tabcolsep}{10pt}
 \centering
    \begin{tabular}{l | l| c | ccccccc}
    \toprule
    \textbf{Method} & \textbf{Backbone}  &  \textbf{B@1} & \textbf{B@2} & \textbf{B@3} &\textbf{ B@4} & \textbf{METEOR} & \textbf{ROUGE} & \textbf{CIDEr} \\
    \hline
    mPLUG-Owl2~\citep{ye2023mplug} & Llama-2-7B & 4.9 &3.2 & 2.3 & 1.7 & 8.7 & 19.6 & 4.5\\
    {Instruct-BLIP~\citep{Dai2023InstructBLIPTG}} & {Vicuna-7B}  & 13.4 & 8.5 & 5.7 & 4.1 & 9.2 & 17.9 & 5.7 \\
    Qwen-VL~\citep{bai2023qwen} & Qwen-7B & 28.8 & 18.4 & 12.2 & 8.6 & 14.1 & 24.3 & 16.6 \\
    {Instruct-BLIP~\citep{Dai2023InstructBLIPTG}} & \multirow{6}{*}{Vicuna-13B}  & 15.9 & 9.9 & 6.4 & 4.4 & 9.6 & 18.6 & 8.0 \\
    {Mini-GPT4~\citep{zhu2023minigpt4}} &  & 31.1 & 16.1 & 8.6 & 5.0 & 11.4 & 20.8 & 6.3 \\
    {Mini-GPT-v2~\citep{chen2023minigpt}} &  & 27.9 & 14.7 & 8.0 & 4.8 & 10.8 & 20.9 & 7.3 \\
    LLaVA \citep{liu2023visual} & & 35.1 & 18.2 & 9.5 & 5.4 & 13.2 & 22.2 & 8.8 \\
    LLaVAR \citep{zhang2023llavar} & & 18.9 & 11.2 & 7.2 & 5.0 & 10.8 & 20.1 & 11.4 \\
    LLaVA 1.5 ~\cite{liu2023improved} & & 31.1 & 16.4 & 9.5 & 6.1 & 11.9 & 21.8 & 15.4 \\
    \hline
    LLaVAR w/ OCR &\multirow{2}{*}{Vicuna-13B} & 21.4 & 13.1 & 8.8 & 6.4 & 12.0 & 22.2 & 13.0 \\
    {LaRA (zero-shot)} & & 29.3 & 19.1 & 12.9 & 9.2 & 14.8 & 26.3 &21.3\\
    \hline
    {LLaVAR (fine-tuned)} &\multirow{2}{*}{Vicuna-13B} & 36.5 & 25.4 & 18.0 & 13.4 & 17.8 & 32.4 & 35.7 \\
    {LaRA} & & \textbf{37.7}  & \textbf{26.4} & \textbf{18.9} & \textbf{14.2} & \textbf{18.4} & \textbf{33.2} & \textbf{46.7} \\
    \bottomrule
    \end{tabular}
    \vspace{-0.5em}
    \caption{Results of different models on text-rich image captioning tasks.}
    \label{table:doc2seq}
    \vspace{-0.9em}
\end{table*}

\subsection{TRINS-Cap: Text-rich image Captioning}
\label{exp:cap}
\vspace{-0.1em}
In our experiments on TRINS-Cap, we ask large multimodal models to generate summaries based on text-rich images. The data set was divided into train, validation, and test sets. We compared LaRA with popular baselines, including InstructBLIP~\citep{Dai2023InstructBLIPTG}, Mini-GPT4~\citep{zhu2023minigpt4}, LLaVA~\citep{liu2023visual}, LLaVAR~\citep{zhang2023llavar}, mPLUG-Owl2~\cite{ye2023mplugowl} and Qwen-VL~\cite{bai2023qwen}. Given that BLIP-2 faced challenges in generating comprehensive and meaningful results for text-rich images, we considered InstructBLIP as an alternative. For all methods, we randomly selected three prompts from ten as instructions for the model (details provided in the Appendix \ref{app:trinscap}). 

Table \ref{table:doc2seq} presents the results of different methods in terms of classical captioning metrics. Models with enhanced visual text understanding generally outperform general multimodal models, such as LLaVA, Mini-GPT4, and Instruct-BLIP. LaRA (zero-shot) refers to the LaRA model fine-tuned on the TRINS-QA dataset and demonstrates improved performance. Comparison of fine-tuned LaRA variants indicates that text recognition ability is still limited for OCR-free methods, suggesting that the CLIP encoder or feature projection process may cause visual information loss. Addressing this limitation may involve employing a better trained encoder on text-rich images or designing a more carefully crafted architecture, a direction we leave for future exploration. 
When fine-tuned with TRINS-Cap, LaRA exhibits much better performance, underscoring the importance of high-quality human-annotated data. Figure~\ref{fig:capexample} shows examples of different models on the TRINS-Cap benchmark. It shows the great capability of LaRA in recognizing text and relating it to its visual contexts, demonstrating the effectiveness of the TRINS dataset. \\

\vspace{-2em}
\subsection{Additional Experiments}
\paragraph{Performance on general visual tasks after TRINS fine-tuning.} We adopted the evaluation protocols of MiniGPT-v2~\citep{chen2023minigpt} and compared LaRA with LLaVA~\cite{liu2023visual} on traditional visual question answering benchmarks in table \ref{tab:visexp}. LaRA shows a comparable performance on knowledgeability and better performance on reasoning and spatial awareness. This further verifies the effectiveness of the TRINS dataset, demonstrating that fine-tuning on text-rich images does not degrade performance on natural images, but instead enhances the results. 
\begin{table}[htp]
\small
\centering
\setlength{\tabcolsep}{6pt} 
\renewcommand{\arraystretch}{1} 
\vspace{-0.3em}
\begin{tabular}{l cccc}
\toprule 
                 &  \textbf{OKVQA}   & \textbf{GQA} & \textbf{VSR} & \textbf{VizWiz}  \\ \midrule
LLaVA~\cite{liu2023visual}           & 57.8  &  41.3 &  {51.2} & 45.0     \\
LaRA                   &     \textbf{58.1}     & \textbf{42.4}        & \textbf{53.0}         & \textbf{53.1}   \\
\bottomrule
\end{tabular}
\vspace{-0.5em}
\caption{Quantitative Results on the public visual benchmarks.}
    \vspace{-1em}
    \label{tab:visexp}
\end{table}
 \begin{table}[htp]
 \centering
\setlength{\tabcolsep}{5pt} 
\small
        \begin{tabular}{ccccccc}
        \toprule
        \textbf{Metrics} &  \textbf{ControlNet} & \textbf{DeepFloyd} & \textbf{TextDiffuser}\\
        \midrule
            FID $(\downarrow)$  & {51.59} & \textbf{49.96} & 51.26   \\
            CLIP Score $(\uparrow)$  &0.3717 & \textbf{0.3917} & 0.3707 \\
            OCR Acc. $(\uparrow)$  &  0.4241 & 0.2192 & \textbf{0.5027}  \\
        \bottomrule
        \end{tabular}
        \vspace{-0.5em}
    \caption{Empirical Results on TRINS-Gen (easy) benchmark.}
    \vspace{-2.2em}
        \label{tab:gen_benchmark1}
\end{table}
\paragraph{Text-to-document generation} 
Diffusion-based text-to-image generation has shown great success, while precise textual renderings remain a big challenge. 
TextDiffuser introduced the MARIO-Eval benchmark, drawing from works such as DrawBench \citep{saharia2022photorealistic} and DrawTextCreative \citep{liu2022character}. However, most text prompts in MARIO-Eval are short and cannot serve as a good evaluation dataset to handle complex real-world human instructions. 
We take advantage of human annotations from TRINS-Cap and build the TRINS-Gen benchmark. It is still difficult to render too many words in a single image~\cite{chen2023textdiffuser}. In response to this, we filter out images with more than 20 OCR words, resulting in a curated set of 2,104 images. We divide these images into two sets (easy and difficult) based on the number of OCR words and the length of the longest OCR string per annotation, where all text prompts in the easy set have less than 9 OCR words. We evaluated existing methods using their public checkpoints and reported the results in Table \ref{tab:gen_benchmark1} and detailed results in Table \ref{tab:gen_benchmark_difficult} (Appendix \ref{app:gebdiff}). 

\begin{figure}[htp]
    \centering
    \includegraphics[width=\linewidth]{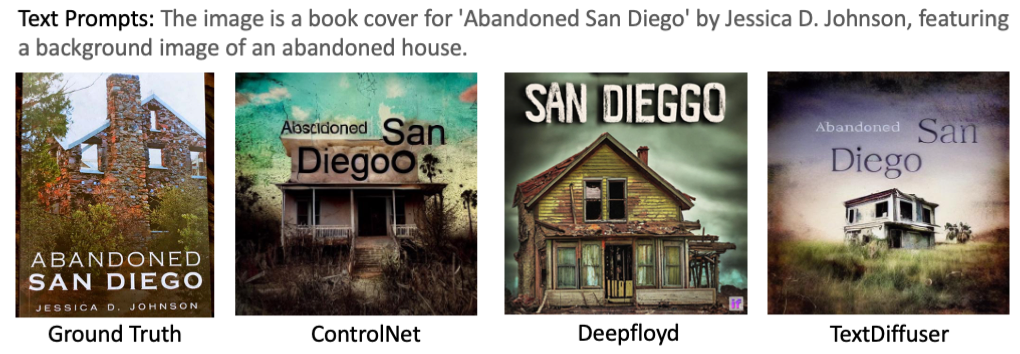}
    \vspace{-1em}
    \caption{Examples generated by different text-to-image models on the TRINS-Gen benchamrk.}
    \vspace{-1em}
    \label{fig:genresults_small}
\end{figure}
\section{Conclusions}

Despite the challenges posed by the prevalence of natural images in training data, 
the significance of visual textual understanding cannot be understated. In this paper, we introduce TRINS, a Text-Rich Image INStruction dataset, comprising a diverse collection of text-rich images, captions, and questions. This dataset, created through a semi-automatic process leveraging large-scale pre-trained models, not only significantly reduces annotation time but also elevates annotation quality.
Furthermore, we propose a novel multimodal language model architecture, LaRA, which incorporates OCR as a pivotal enhancement for text-rich image understanding. We anticipate that continued progress in multimodal language model architectures, fine-tuning techniques, and the expansion of diverse, text-rich datasets like TRINS will push the boundaries of visual textual understanding. This, in turn, will facilitate more efficient collaboration between humans and agents, potentially revolutionizing numerous real-world applications.

{
    \small
    \bibliographystyle{ieeenat_fullname}
    \bibliography{main}
}

\appendix
\newpage

~\newpage
\section{Data Annotation}
\label{app:datadetail}
\begin{table}[H]
\small
\centering
\begingroup
\setlength{\tabcolsep}{7pt} 
\renewcommand{\arraystretch}{1} 
\begin{tabular}{lcc}
\toprule
Number of images & 39,153 \\
Number of human annotations & 39,153 \\
Average \# of words per human annotations & 65.1 \\
Number of QA pairs & 102,437 \\
Average \# of words per question & 10.5\\
Average \# of words per answer & 23.9\\
\bottomrule
\end{tabular}
\endgroup
\vspace{-1em}
\caption{TRINS datasets summary}
\vspace{-1em}
\label{tab: stats}
\end{table}

\begin{figure*}[t]
    \centering
    \hspace{-4mm}
    \includegraphics[width=0.92\textwidth]{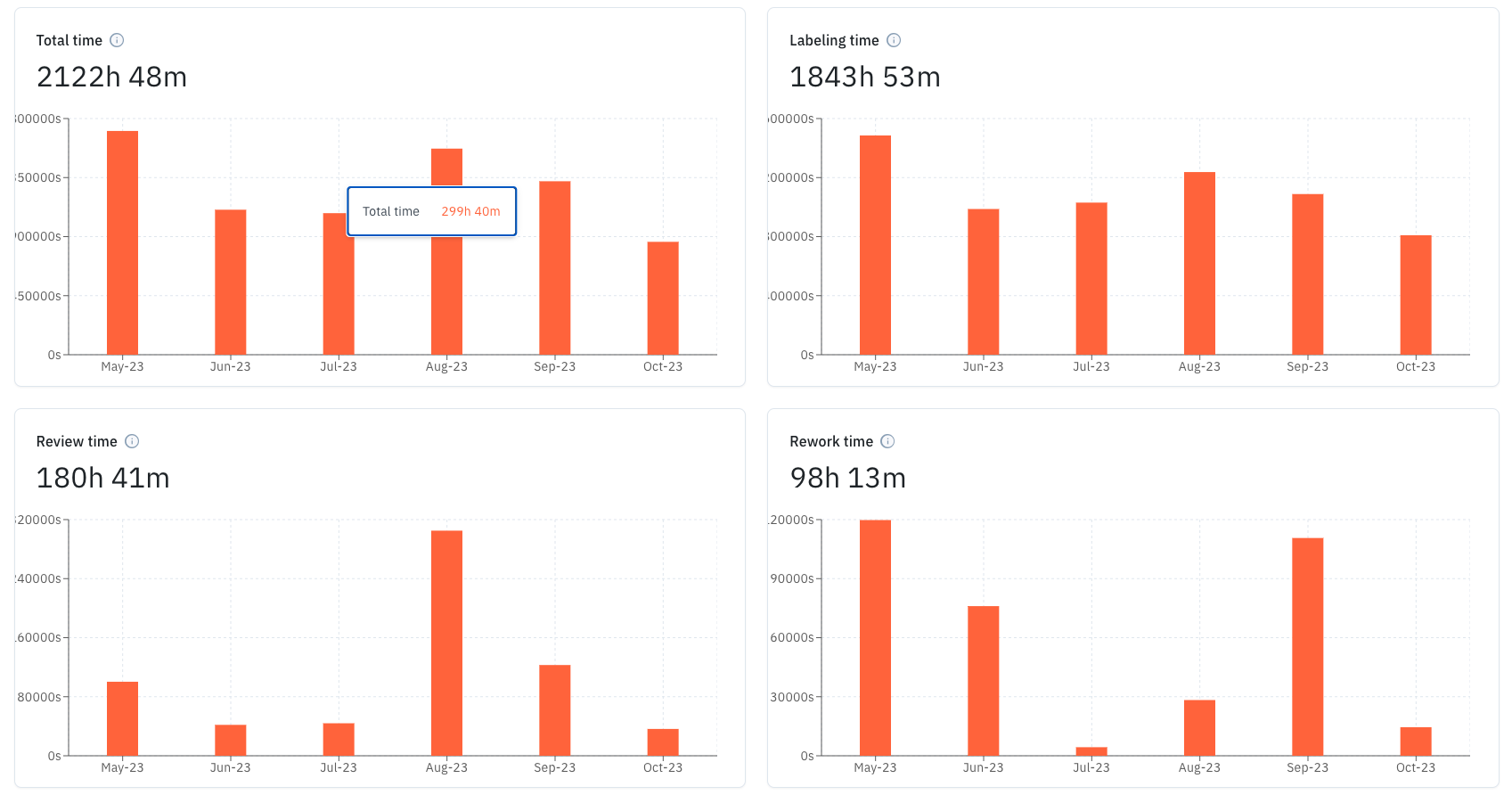}
    \caption{TRINS datasets annotation statistics}
    \label{fig:annotationstats}
\end{figure*}

\begin{figure*}[h!]
    \centering
    \hspace{-4mm}
    \includegraphics[width=\textwidth]{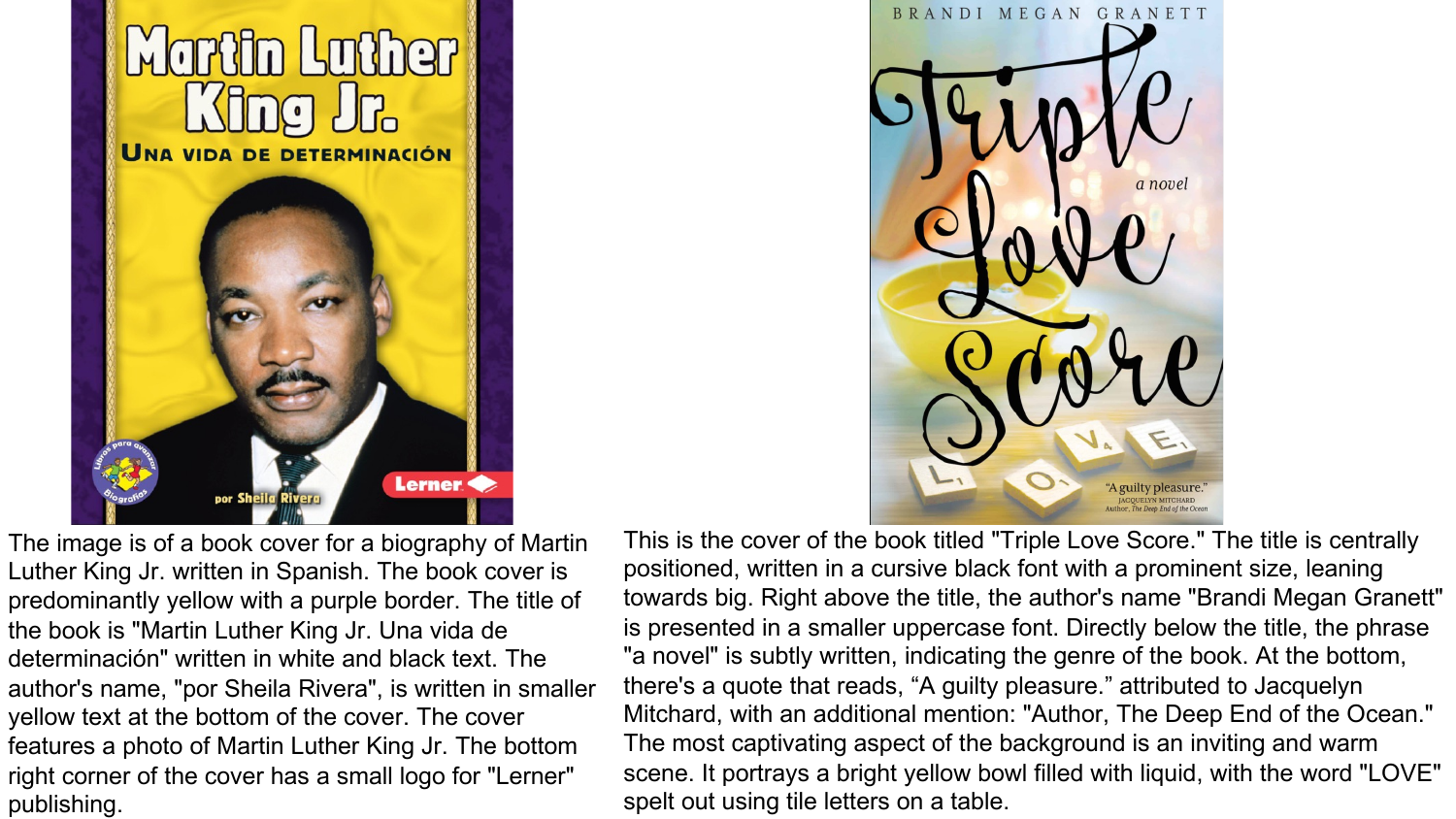}
    \caption{TRINS-Cap Human Annotated Examples}
    \label{fig:cap3}
\end{figure*}

\begin{figure*}[h!]
    \centering
    \hspace{-4mm}
    \includegraphics[width=\textwidth]{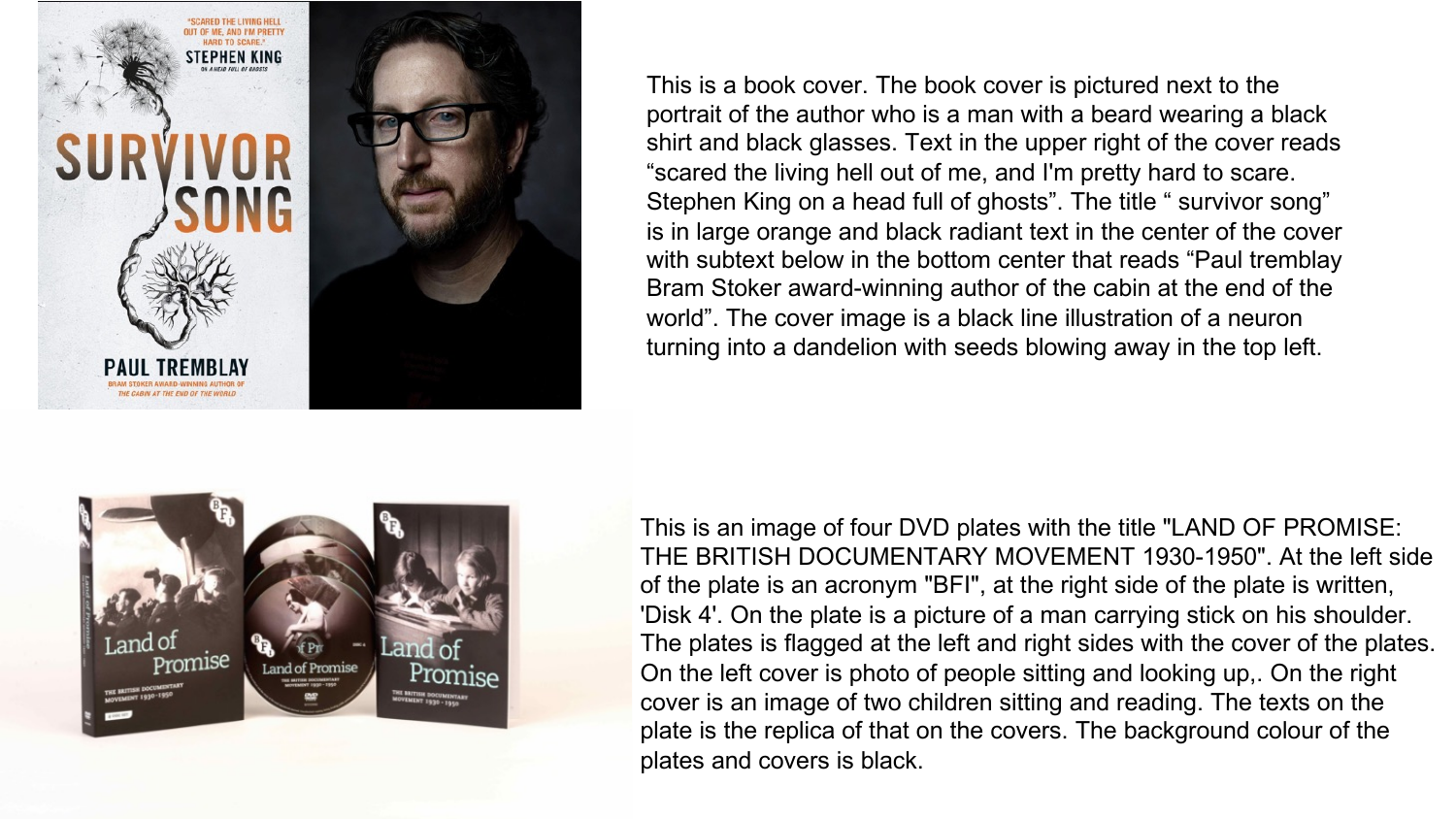}
    \includegraphics[width=\textwidth]{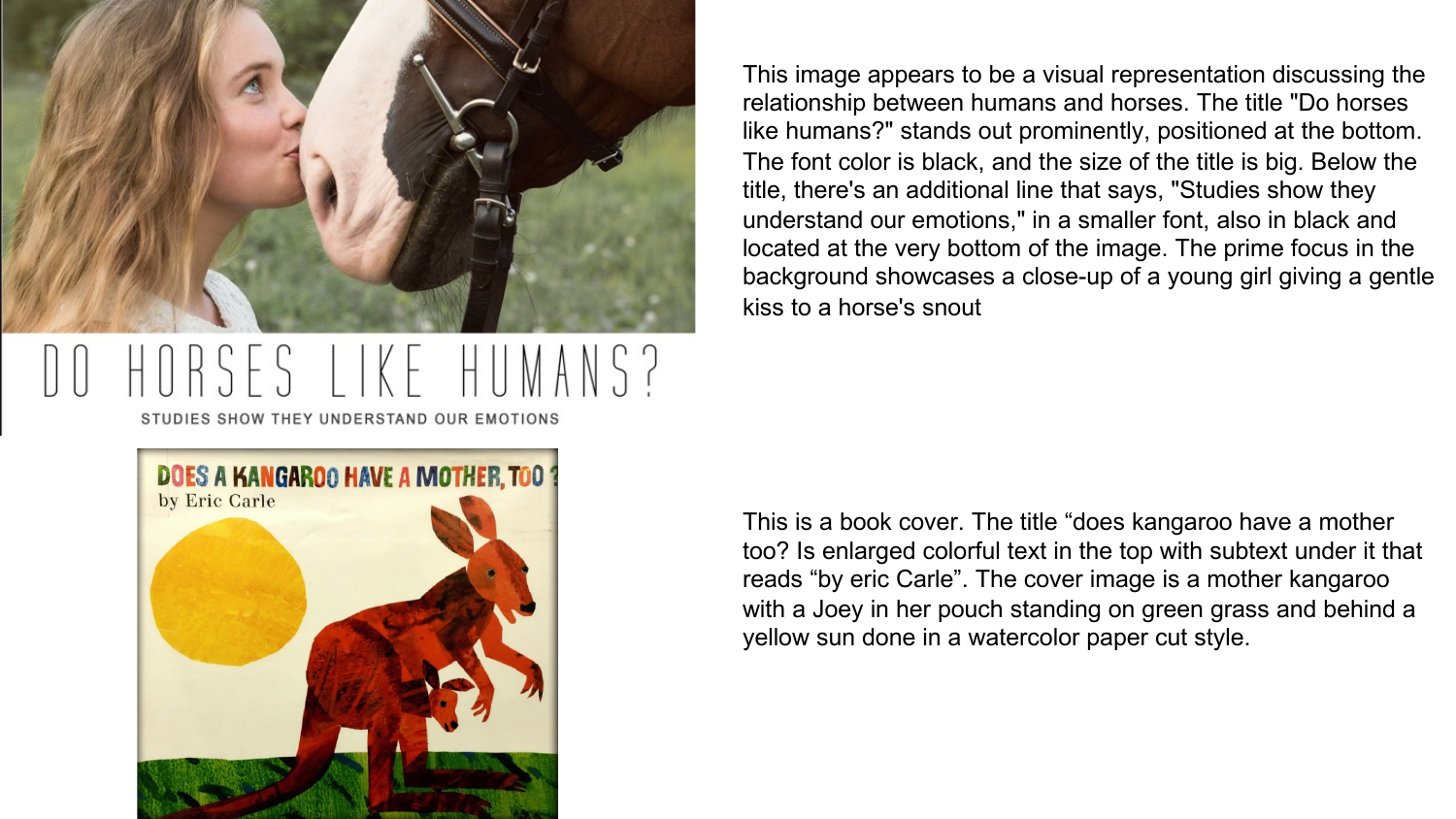}
    \caption{TRINS-Cap Human Annotated Examples}
    \label{fig:cap1}
\end{figure*}

\begin{figure*}[h!]
    \centering
    \hspace{-4mm}
    \includegraphics[width=\textwidth]{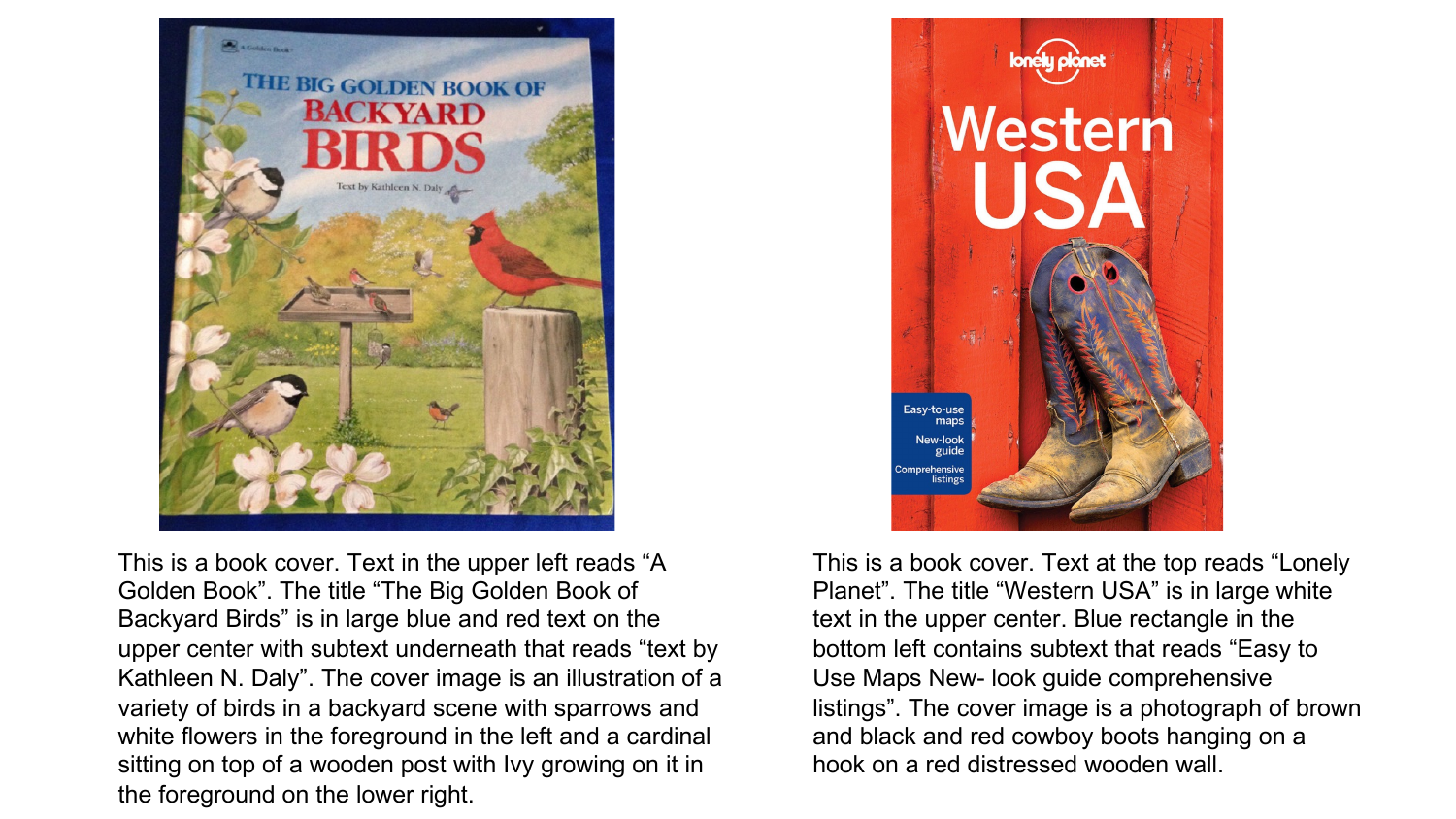}
    \includegraphics[width=\textwidth]{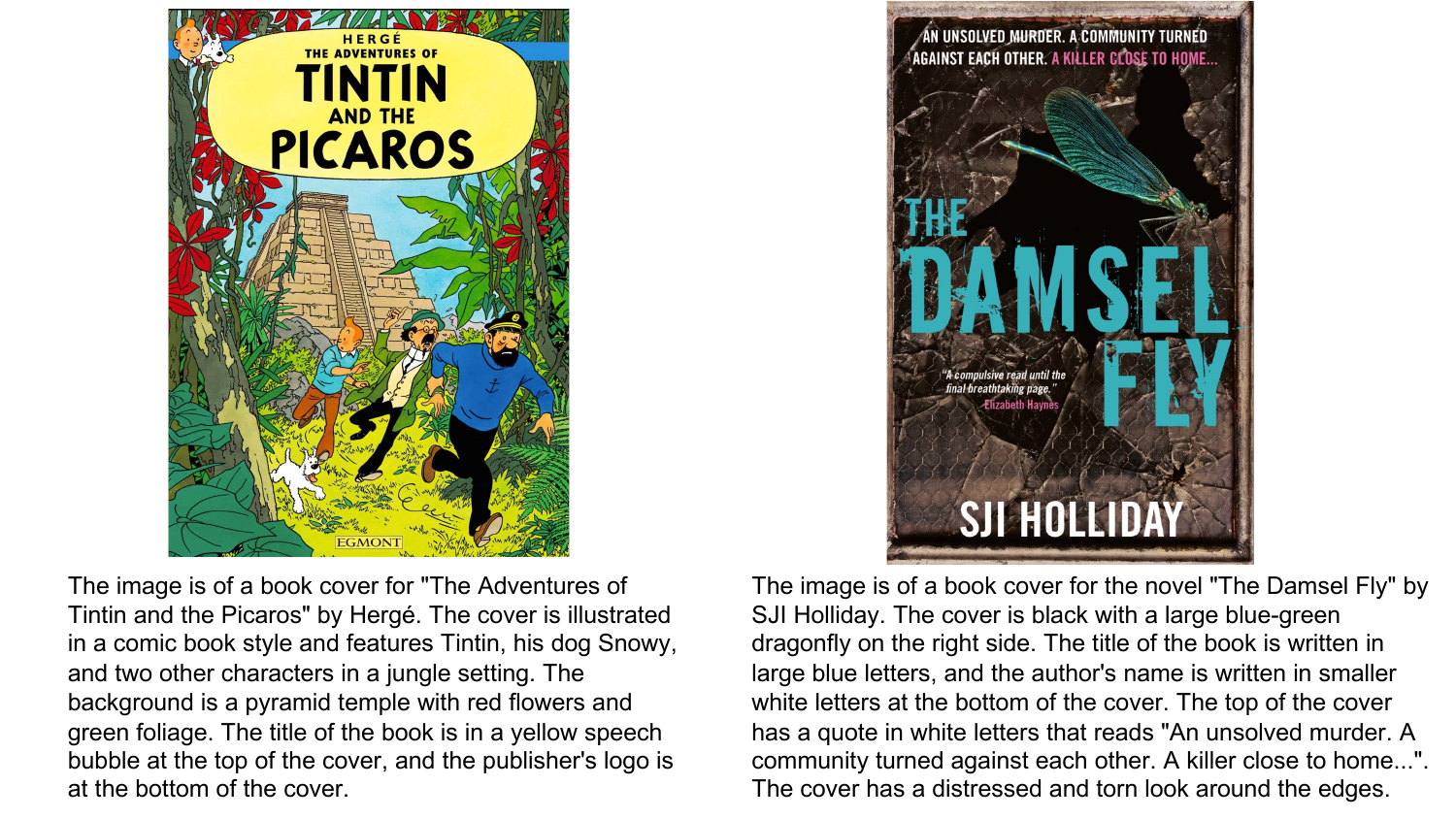}
    \caption{TRINS-Cap Human Annotated Examples}
    \label{fig:cap2}
\end{figure*}

\begin{figure*}[h!]
    \centering
    \hspace{-4mm}
    \includegraphics[width=\textwidth]{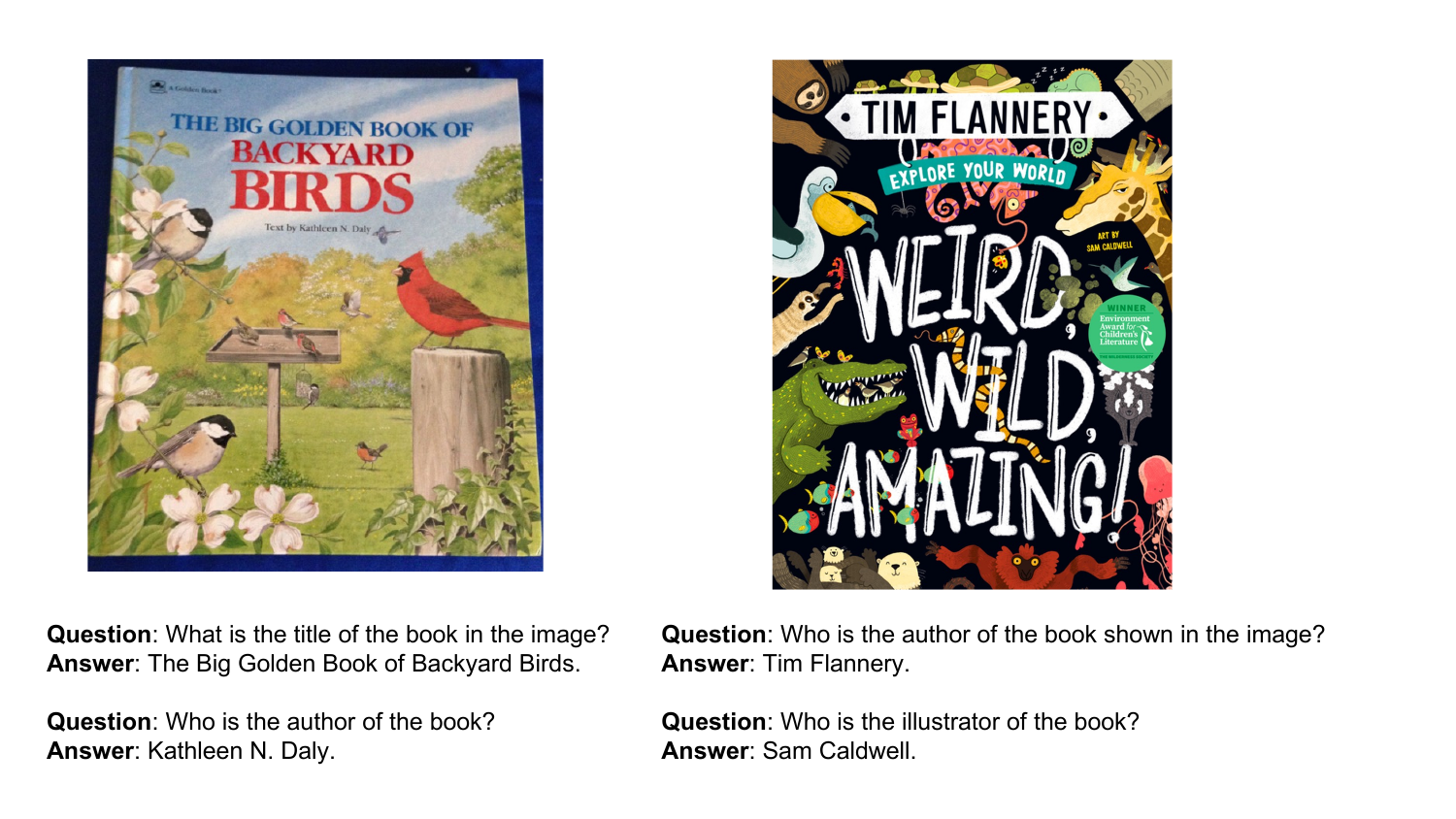}
    \includegraphics[width=\textwidth]{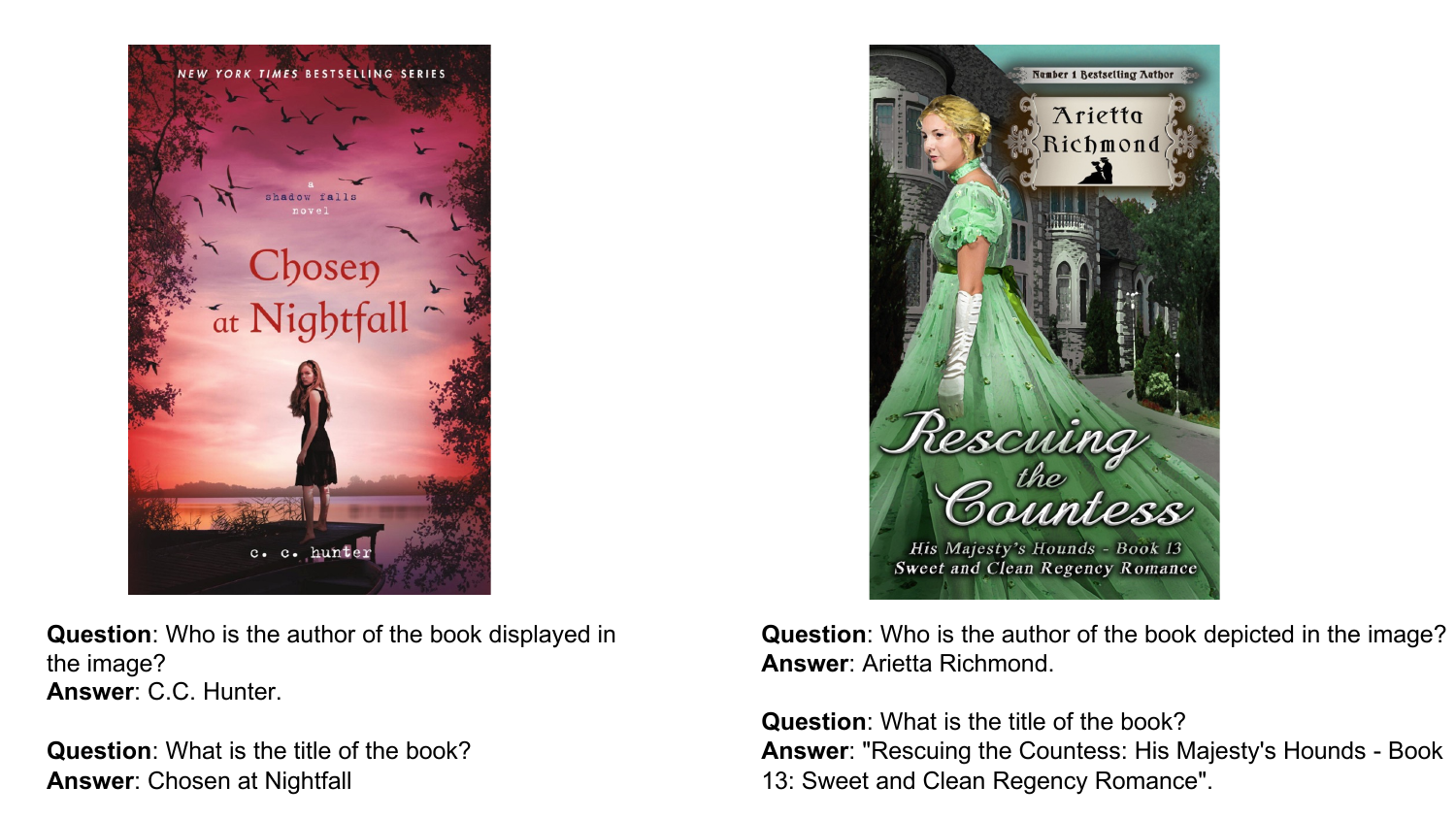}
    \caption{TRINS-QA Human-assisted Annotated Examples}
    \label{fig:qa1}
\end{figure*}

\section{TRINS-Gen: Text-to-Image Generation}
\paragraph{Text-Rich Image Generation} Diffusion-based Text-to-image Generation has shown great success, while precise textual renderings remain a big challenge. Imagen \citep{saharia2022photorealistic}, eDiff-I \citep{balaji2022ediffi}, and DeepFolyd \citep{DeepFloyd} exploit large T5 models \citep{raffel2020exploring} for better precision. The study by Liu et al. \citep{liu2022character} proposes a character-sensitive alternative to traditional token length handling. GlyphDraw \citep{ma2023glyphdraw} merges superior images with Chinese texts, while Textdiffuser \citep{chen2023textdiffuser} uses the Transformer model for multiline text generation and segmentation masks for user control. However, manual input is still needed for key-term recognition. TextDiffuser also introduced the MARIO-Eval benchmark, drawing from works such as DrawBench \citep{saharia2022photorealistic} and DrawTextCreative \citep{liu2022character}. However, most text prompts in MARIO-Eval are short and cannot serve as a good evaluation dataset to handle complex real-world human instructions. In TRINS-Gen, we introduce a new human-annotated image generation benchmark for existing text-rendering methods, together with a new training dataset.

\begin{figure*}[htp]
    \centering
    \hspace{-4mm}
    \includegraphics[width=0.92\textwidth]{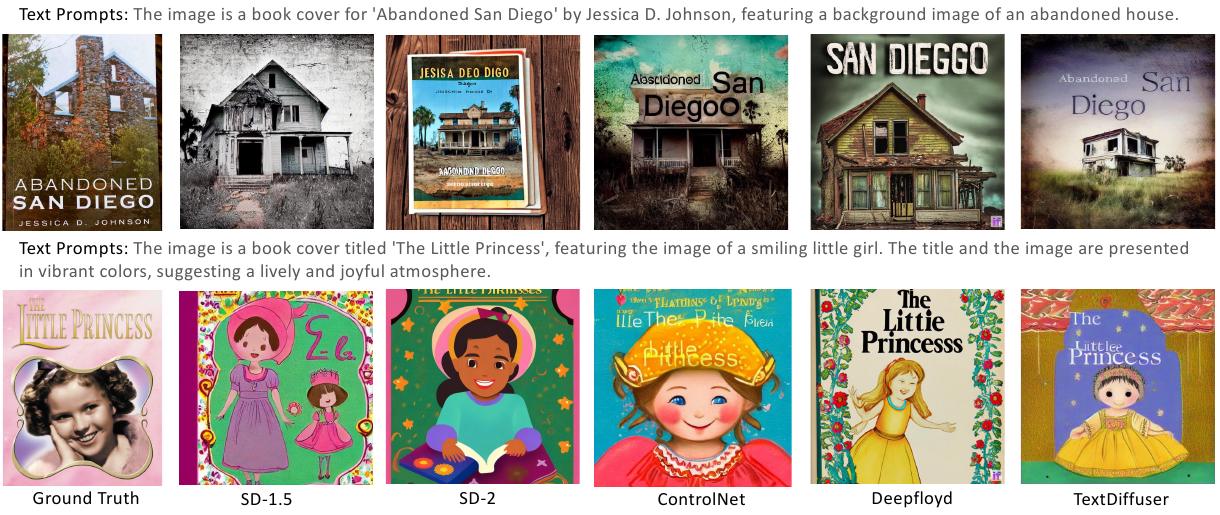}
    \vspace{-1em}
    \caption{Examples generated by different text-to-image models on the TRINS-Gen benchamrk.}
    \label{fig:genresults}
\end{figure*}

Text rendering is a main challenge for text-rich image generation, and we do an extensive evaluation of existing methods on the TRINS-Gen benchmark. We take advantage of human annotations and build the TRINS-Gen benchmark.  We have chosen data examples with fewer words, as existing methods cannot yet render too many texts within a single image. Following \citet{chen2023textdiffuser}, we consider the calculated FID score between the generated images and the TRINS-Gen ground truth images. In addition, we calculate the CLIP scores to measure whether the visual components within the generated image followed the text instructions, and OCR accuracy to measure whether the desired texts are generated correctly. We evaluated existing methods using their public checkpoints and reported the results in Table \ref{tab:gen_benchmark}.  We can see that DeepfFloyd has the best image quality and aligns well with the user-text prompts. ControlNet, which takes a rendered black-white text image as a condition, has an image quality similar to that of TextDiffuser. In terms of text rendering ability, TextDiffuser is best compared to other baselines. In Figure \ref{fig:genresults}, we show two examples of images generated by different methods. Stable Diffusion (SD) models cannot render any meaningful text. Deepfloyd and ControlNet may miss some words or render texts with spelling errors. TextDiffuser can generally render text well when the length of words is short. However, all methods do not render the desired texts when the number of words increases, as described in the Appendix \ref{app:gebdiff}. The main cause in TextDiffuser~\cite{chen2023textdiffuser} is that its layout models cannot parse complex prompts, and an LLM-based layout model may help.
 \begin{table*}[htbp]
 \centering
\setlength{\tabcolsep}{5pt} 
\small
        \begin{tabular}{ccccccc}
        \toprule
        \textbf{Metrics} & \textbf{SD-2.0} & \textbf{SD-1.5} & \textbf{ControlNet} & \textbf{DeepFloyd} & \textbf{TextDiffuser}\\
        \midrule
            FID $(\downarrow)$ & 59.54 & 59.05 & 51.59 & 49.96 & 51.26   \\
            CLIP Score $(\uparrow)$ & 0.3520 & 0.3419 &0.3717 & 0.3917 & 0.3707 \\
            OCR Accuracy $(\uparrow)$ & 0.0893 &  0.2205 &  0.4241 & 0.2192 & 0.5027  \\
            OCR Precision $(\uparrow)$ & 0.1019 &  0.2535 & 0.4083 & 0.2225 & 0.6001  \\
            OCR Recall $(\uparrow)$ &  0.0900 &  0.2218 &  0.4268  & 0.2669 & 0.5061  \\
            OCR F-measure $(\uparrow)$ & 0.0956 &  0.2366 & 0.4174 &  0.2427 & 0.5491 \\
        \bottomrule
        \end{tabular}
        \vspace{-1em}
    \caption{Empirical Results of different methods on TRINS-Gen (easy) benchmark.}
        \label{tab:gen_benchmark2}
\end{table*}

\subsection{DocGen: Text-rich Image Generation}
\label{dataset:gen}
Given high-quality text-rich image descriptions, it is natural to consider their application in training and evaluating text-rich image generation models~\cite{chen2023textdiffuser}. As highlighted in Section~\ref{dataset:cap}, the average number of OCR words is 65.1. However, generating such a substantial amount of text within a single image poses a significant challenge for diffusion models. In response to this, we have established the first human-annotated text-rich image generation benchmark. This benchmark involves filtering out images with more than 20 OCR words, resulting in a curated set of 2,104 images. We divide these images into two sets based on the number of OCR words and the length of the longest OCR string per annotation, where all text prompts in the easy set have less than 9 OCR words with the longest OCR string of 5. Visual examples of both data sets are included in Appendix \ref{app:datadetail}.

\section{Additional Experiments}
\subsection{More TRINS-Cap Results}
\begin{figure*}[h!]
    \centering
    \hspace{-4mm}
    \includegraphics[width=\textwidth]{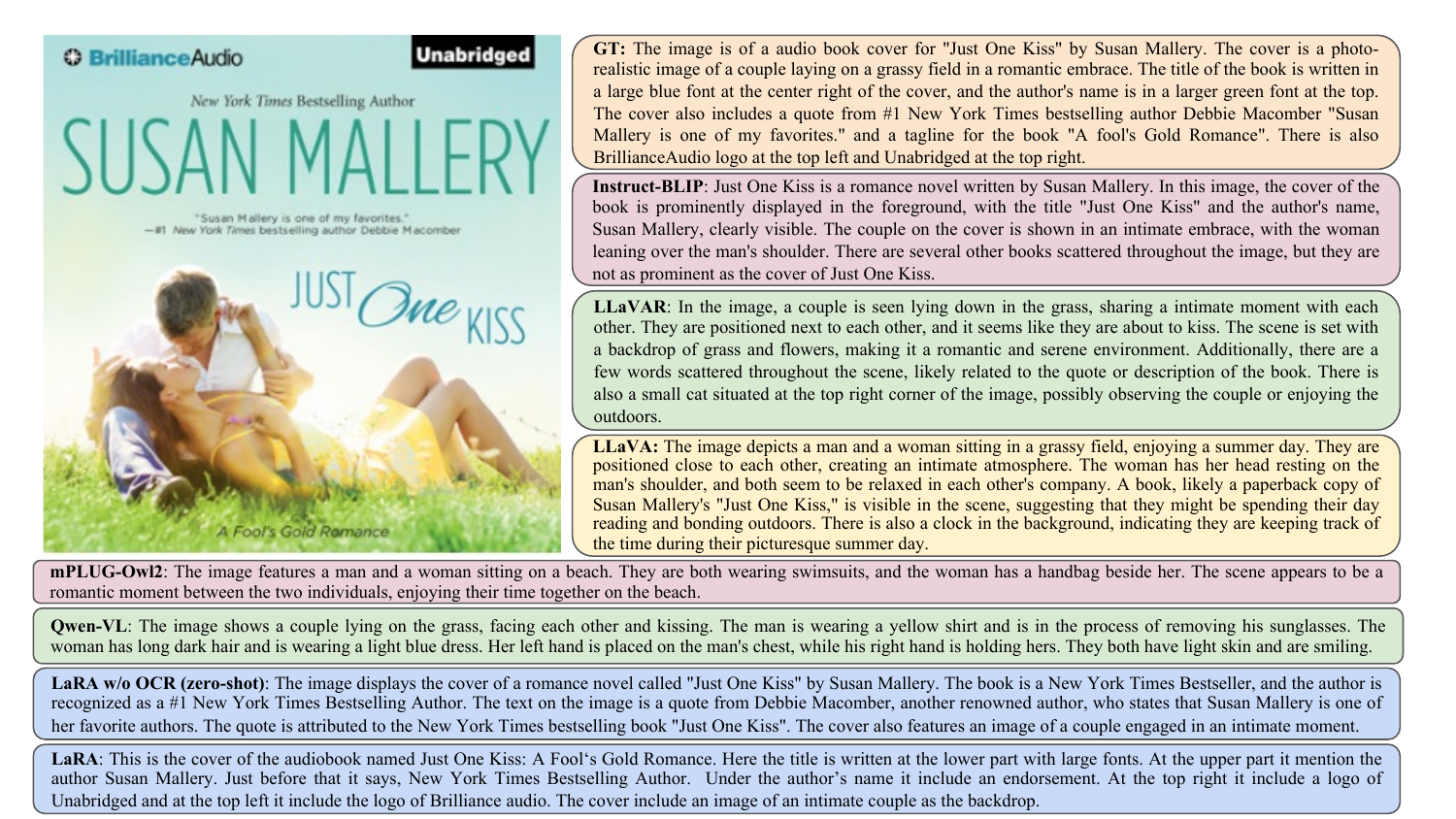}
    \includegraphics[width=\textwidth]{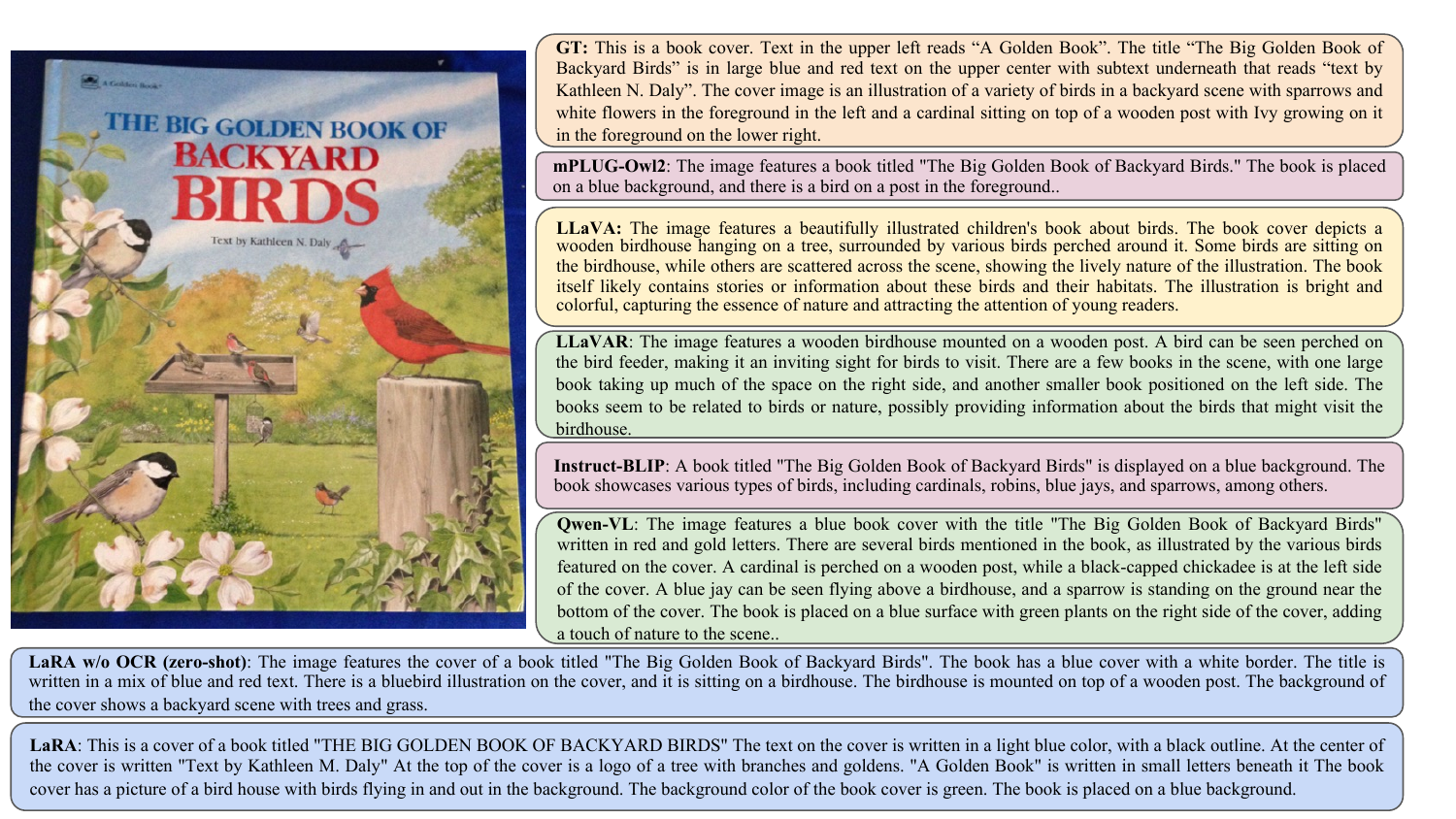}
    \caption{More Generated Examples by different methods on TRINS-Cap.}
    \label{fig:cap_results}
\end{figure*}

Figure \ref{fig:cap_results} shows more results of random examples from TRINS-Cap.
\label{app:trinscap}

\subsection{More TRINS-QA Results}
\label{app:trinsqa}

\begin{figure*}[h!]
    \centering
    \hspace{-4mm}
    \includegraphics[width=\textwidth]{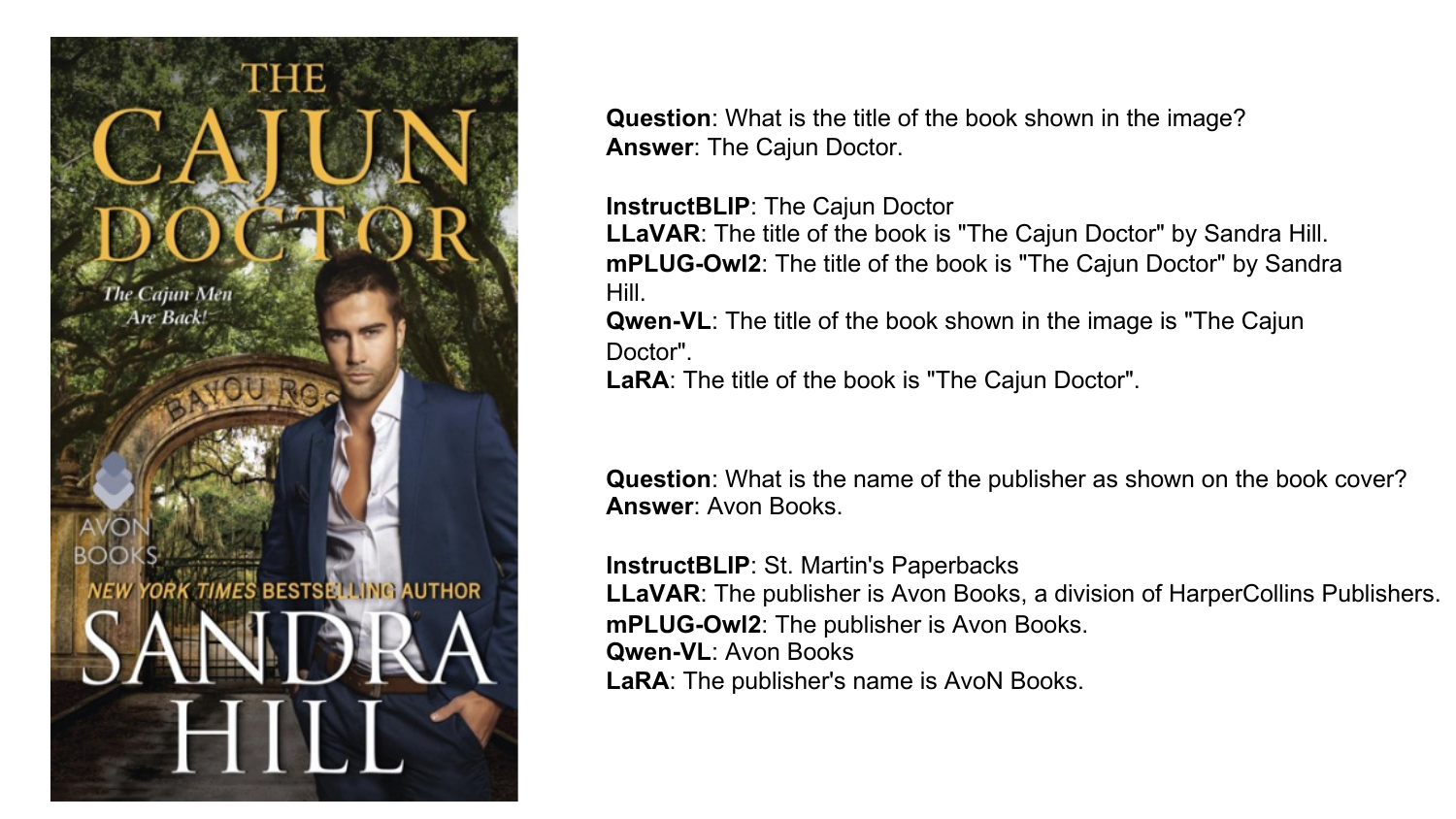}
    \includegraphics[width=\textwidth]{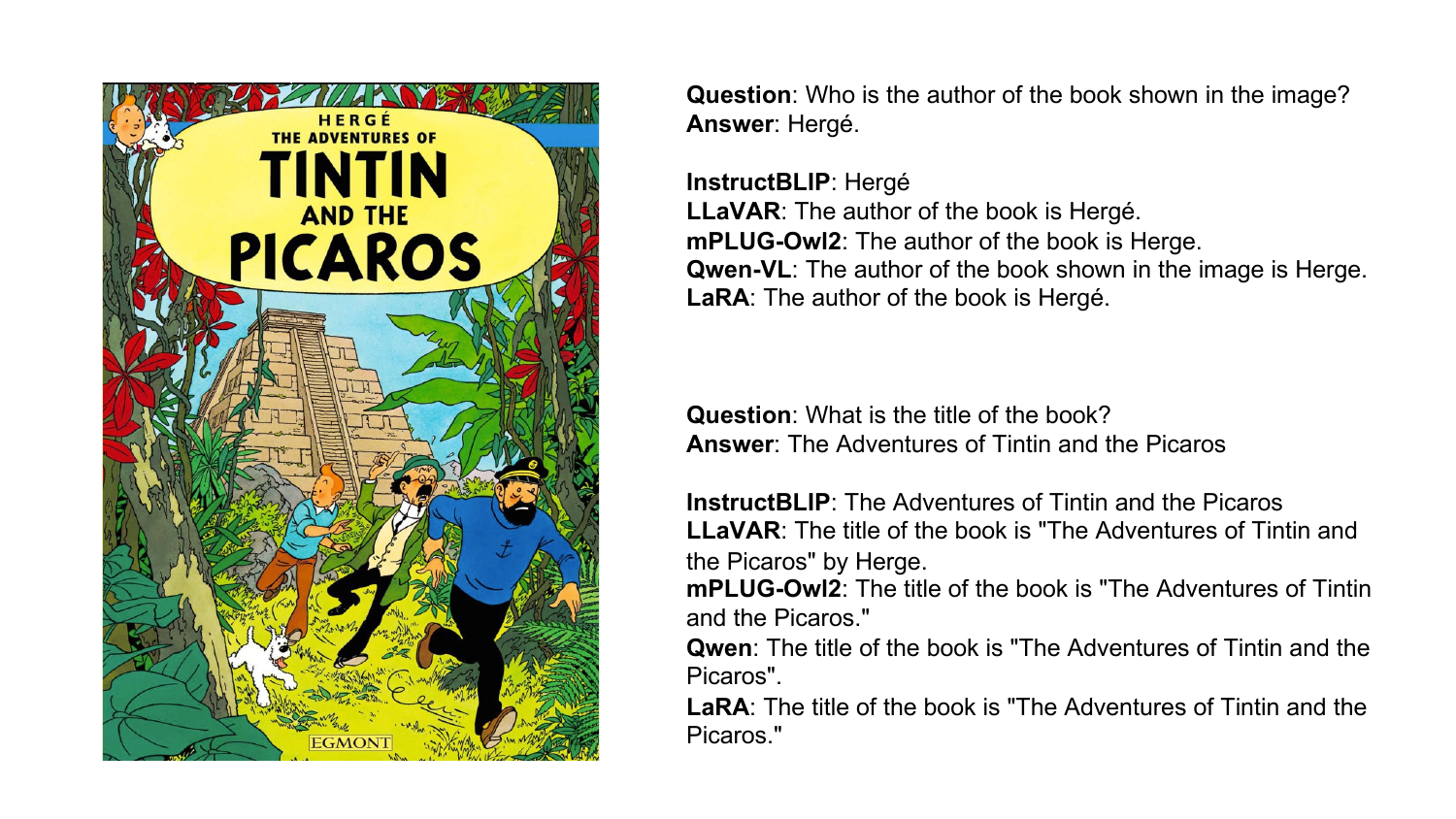}
    \caption{TRINS-QA Extract QA Results}
    \label{fig:aqamore}
\end{figure*}

\begin{figure*}[h!]
    \centering
    \hspace{-4mm}
    \includegraphics[width=\textwidth]{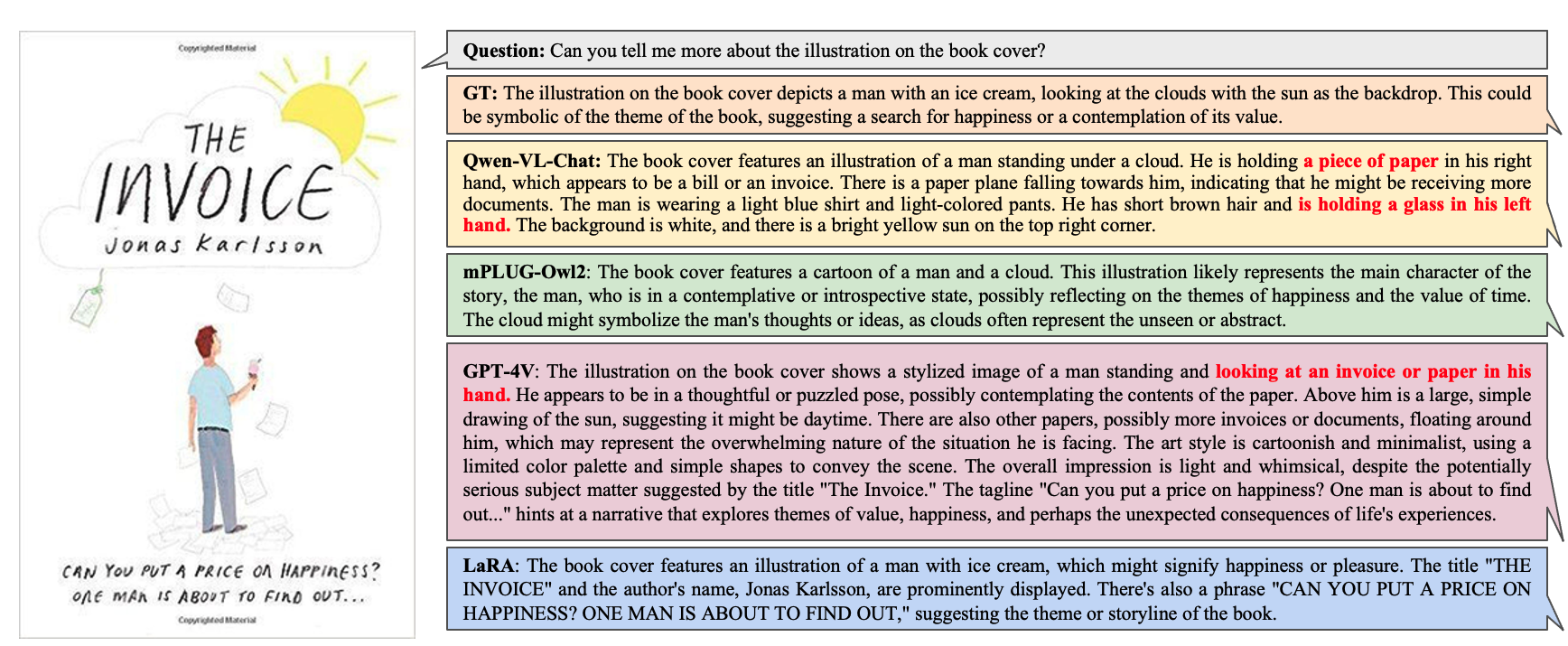}
    \includegraphics[width=\textwidth]{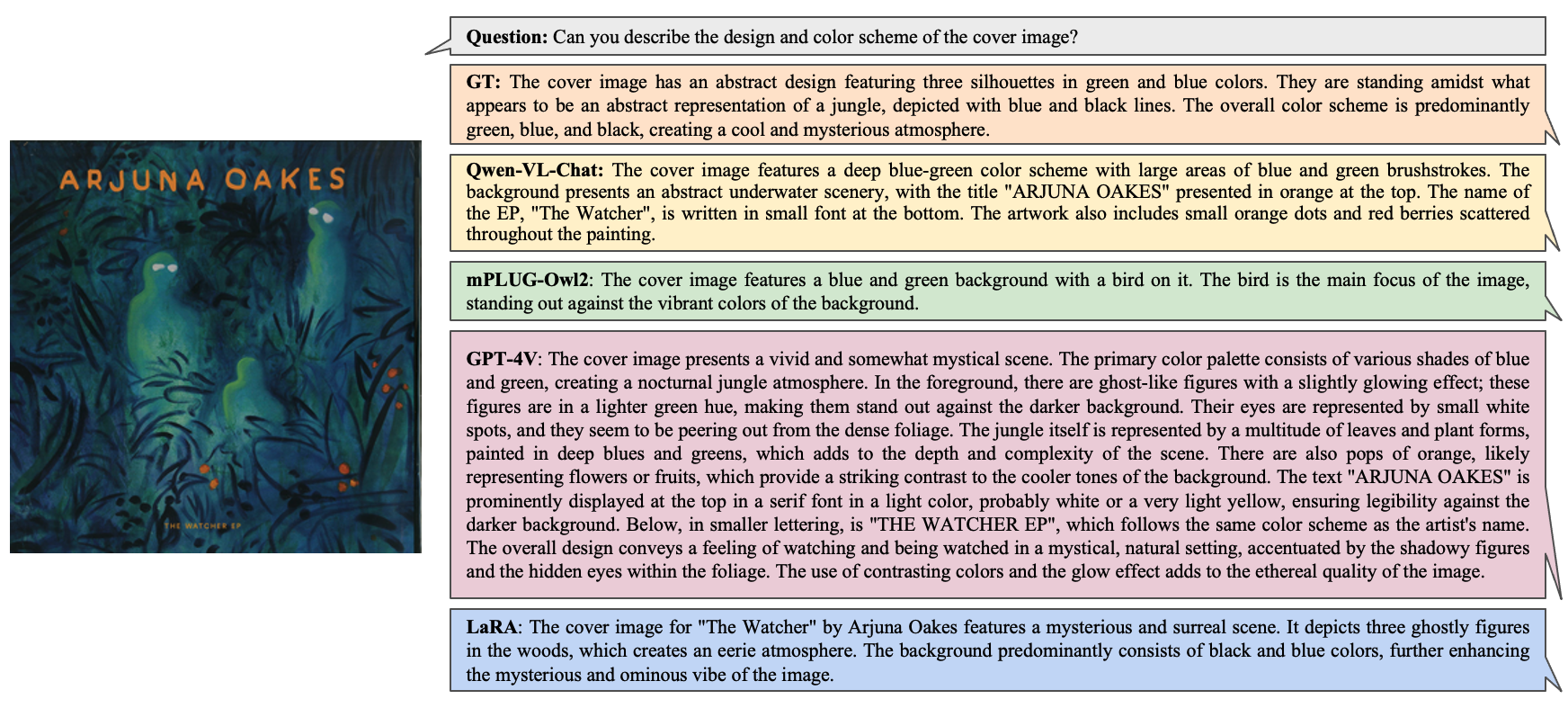}
    \caption{TRINS-QA Abstract QA Results}
    \label{fig:qamore}
\end{figure*}

Figure \ref{fig:qamore} shows more random examples of the extract QA task and Figure \ref{fig:aqamore} shows two more random examples of the abstract QA task. 

\subsection{More TRINS-Gen Results}
\label{app:gebdiff}
 \begin{table}[htbp]
 \centering
\setlength{\tabcolsep}{2pt} 
\small
        \begin{tabular}{ccccccc}
        \toprule
        \textbf{Metrics} & \textbf{SD-2.0} & \textbf{SD-1.5} & \textbf{ControlNet} & \textbf{TextDiffuser}\\
        \midrule
            FID $(\downarrow)$ & 55.48 & 56.73 & 60.15 & 74.56   \\
            CLIP Score $(\uparrow)$ & 0.3512 & 0.3384 &0.3301  & 0.3219 \\
            OCR Accuracy $(\uparrow)$ &  0.0447 & 0.0256 &  0.0833  & 0.0952  \\
            OCR Precision $(\uparrow)$ &  0.0877 & 0.0513  & 0.1011 & 0.1249  \\
            OCR Recall $(\uparrow)$ &  0.0450 &  0.0252  &  0.0827  & 0.0941  \\
            OCR F-measure $(\uparrow)$ &  0.595 & 0.0338  & 0.910 & 0.1073 \\
        \bottomrule
        \end{tabular}
        \vspace{-1em}
    \caption{Empirical Results of different methods on TRINS-Gen difficult benchmark.}
        \label{tab:gen_benchmark_difficult}
\end{table}

\begin{table*}[t]
\small
\centering
\begingroup
\setlength{\tabcolsep}{5pt} 
\renewcommand{\arraystretch}{1} 
\begin{tabular}{l cccccccc}
\toprule 
                   &  \textbf{ST-VQA} & \textbf{OCR-VQA} & \textbf{TextVQA} & \textbf{DocVQA} & \textbf{ChartQA} & \textbf{InfoVQA} & \textbf{FUNSD} & \textbf{SROIE}\\ \midrule
BLIP-2 \citep{li2023blip2} $\dagger$                  & 21.7            & 30.7             & 32.2             & 4.9        & 3.4 & 11.3 & 0.20 & 0.14     \\
OpenFlamingo \citep{anas_awadalla_2023_7733589} $\dagger$                      & 19.3            & 27.8             & 29.1             & 5.1   & 9.1 & 15.0 & 0.85 & 0.12          \\
MiniGPT4 \citep{zhu2023minigpt4} $\dagger$                    & 14.0              & 11.5             & 18.7             & 3.0  & 4.3 & 13.3 & 1.19 & 0.04             \\
mPLUG-Owl \citep{ye2023mplugowl} $\dagger$                            & 29.3            & 28.6             & 40.3             & 6.9 &9.5 & 16.5 & 1.02 & 0.60            \\ 
LLaVA \citep{liu2023visual} $\dagger$                 & 28.9            & 11.0               & 36.7             & 6.9 &    28.9 & 13.8 & 1.02 & 0.12\\
LLaVA1.5 \citep{liu2023visual} $\dagger$                 & 38.1            & 58.1               & 38.7             & 8.5   & 9.3 & 14.7 & 0.20 & 1.70  \\
LLaVAR $\dagger$               & 39.2 & 23.8 & 48.5 & 11.6 & 12.2 & 16.5 & 0.50 & 5.20    \\
mPLUG-Owl2 \citep{ye2023mplug} $\dagger$                            & 29.3            & 28.6             & 40.3             & 6.9  & 19.4 & 18.9 & 1.40 & 3.20                  \\
\midrule
LLaVAR (finetuned)                          & 40.3          & 28.8          & 50.1          & 12.2 & - & - & - & -\\ 
LLaVAR w/ OCR                            & 44.8          & 42.7        & 60.4         & 48.3 & - & - & - & -\\ 
LaRA                           & 47.2          & 41.2        & 59.9         & 50.8 & 25.6 & 28.4 & 23.2 & 36.6 \\
\bottomrule
\end{tabular}
\endgroup
\caption{\label{table: VQA result} Zero-shot performance (accuracy \%) on text-based VQA. We use $\dagger$ to refer to the results obtained from \citet{liu2023hidden}.} 
\label{tab:vqa_result}
\vspace{-2em}
\end{table*}

Figure \ref{tab:gen_benchmark_difficult} shows the results of different methods on the TRINS-Gen difficult benchmark. Deeploy model needs more than two A100-80GB GPUs and we omit its result due to the resource limitation. SD-2.0 model demonstrated the best performance in image quality and alignment with text descriptions, outperforming SD-1.5, ControlNet, and TextDiffuser. TextDiffuser, however, showed its strength in accurately generating text within images, leading in metrics related to text recognition. ControlNet exhibited balanced performance in these text-related tasks, making it a notable contender in terms of precision and recall balance. In general, when text prompts become complex, SD-2.0 and SD-1.5 can still generate meaningful images related to the prompts. Textdiffuser struggle to place to many words in the image and fail to generate good images. In addition, its text generation quality is much lower as the number of generated words increases. Based on the results, it is obvious to see that TRINS-Gen difficult evaluation set is more challenging and need to be better tackled by future methods.

\begin{table*}[h!]
\small
\setlength{\tabcolsep}{7pt}
\centering
\begin{tabular}{l | c| c | ccccccc}
\toprule
\multirow{2}{*}{\textbf{Method}}  & \multirow{2}{*}{\textbf{Resolution}} & \textbf{Extract} &\multicolumn{7}{c|}{\textbf{Abstract}} \\
\cline{4-10}
 &   & \textbf{Accuracy} & \textbf{B@1} & \textbf{B@2} & \textbf{B@3} &\textbf{ B@4} & \textbf{METEOR} & \textbf{ROUGE} & \textbf{CIDEr} \\
\hline
LLaVA \citep{liu2023visual} & \multirow{4}{*}{$336^2$} & 22.6 & 12.3 & 7.5 & 4.9 & 3.4 & 15.3 & 20.2 & 26.2 \\
LLaVAR \citep{zhang2023llavar} & &44.1 &  27.5 & 19.5 & 14.4 & 10.9 & 21.6 & 33.0 & 94.7 \\
{LLaVAR (finetuned)} &  & 50.1  & 26.1 & 18.7 & 14.1 & 11.0 & 23.1 & 33.4 & 112.6 \\
LaRA & & 58.8 & 31.0 & 23.6 & 18.6 & 15.1 & 26.5 & 38.0 & 135.6 \\
\bottomrule
\end{tabular}
\caption{Results of different models on TRINS-VQA-Human for text-rich image question-answering tasks.}\vspace{-0.5em}
\label{tab:TRINS-VQA2}
\end{table*}

\section{More Annotation Examples}
Figure \ref{fig:annotationstats} shows the annotation statistics from the LabelBox. 
\label{app:moreexamples}

\end{document}